\documentclass[drones,article,submit,moreauthors,pdftex]{mdpi}
\firstpage{1} 
\pubvolume{1}
\issuenum{1}
\articlenumber{0}
\pubyear{2022}
\copyrightyear{2022}
\datereceived{} 
\dateaccepted{} 
\datepublished{} 
\hreflink{https://doi.org/} 

\usepackage{xcolor}
\usepackage{subcaption}

\Title{Drone Detection and Tracking in Real-Time by Fusion of Different Sensing Modalities}

\TitleCitation{Title}

\Author{Fredrik Svanström $^{1,\dagger}$, Fernando Alonso-Fernandez $^{2,}$*\orcidA{0000-0002-1400-346X} and Cristofer Englund $^{2,3,\ddagger}$}

\AuthorNames{Fredrik Svanström, Fernando Alonso-Fernandez and Cristofer Englund}

\AuthorCitation{Lastname, F.; Lastname, F.; Lastname, F.}

\address{%
$^{1}$ \quad Air Defence Regiment. Swedish Armed Forces\\
$^{2}$ \quad School of Information Technology, Halmstad University, SE 301 18 Halmstad, Sweden\\
$^{3}$ \quad RISE, Lindholmspiren 3A, SE 417 56 Gothenburg, Sweden}

\corres{Correspondence: feralo@hh.se}

\firstnote{Email: DroneDetectionThesis@gmail.com} 
\secondnote{Email: cristofer.englund@hh.se}

\abstract{
Automatic detection of flying drones is a key issue where its presence, especially if unauthorized, can create risky situations or compromise security.
Here, we design and evaluate a multi-sensor drone detection system.
In conjunction with standard video cameras and microphone sensors, we explore the use of thermal infrared cameras, pointed out as a feasible and promising solution that is scarcely addressed in the related literature.
Our solution integrates a fish-eye camera as well to monitor a wider part of the sky and steer the other cameras towards objects of interest.
The sensing solutions are complemented with an ADS-B receiver, a GPS receiver, and a radar module. However, our final deployment has not included the latter due to its limited detection range.
The thermal camera is shown to be a feasible solution as good as the video camera, even if the camera employed here has a lower resolution. 
Two other novelties of our work are the creation of a new public dataset of multi-sensor annotated data that expands the number of classes compared to existing ones, as well as the study of the detector performance as a function of the sensor-to-target distance.
Sensor fusion is also explored, showing that the system can be made more robust in this way, mitigating false detections of the individual sensors.
}


\begin{document}



\section{Introduction}

Drones, also referred to small and remotely controlled unmanned aerial vehicles (UAVs), can fulfil valuable societal roles such as law enforcement, medical, construction, search and rescue, parcel delivery, remove area exploration, topographic mapping, forest/water management, or inspection of big infrastructures like power grids \cite{drones6060147}. 
%
%
Their low cost and ease of operation have caused drones to find their way into consumer use just for recreation and entertainment as well \cite{[FAI]}.
Unfortunately, they can also be intentionally or unintentionally misused, threatening the safety of others. 
For example, an aircraft can be severely damaged if it collides with a consumer-sized drone, even at moderate speeds 
\cite{[Dayton]}, and an ingested drone can rapidly disable an aircraft engine.
An increasingly common risk is the report of drone sightings in restricted airport areas, which ultimately has led to total closure of the airport and cancellation of hundreds of flights \cite{UAVincidents}. 
Several near-misses and verified collisions with UAVs have involved hobbyist drone operators violating aviation safety regulations, sometimes without knowledge.
This rapid development in the use and potential misuse of UAVs has consequently produced an increase in research on drone detection \cite{Taha19,[GoogleTrends]} to counteract potential risks due to intrusion in restricted areas, either intentional or unintentional. 

In this work, we address the design and evaluation of an automatic multi-sensor drone detection and tracking system.
Our developments are built on state-of-the-art machine learning techniques, extending methods from conclusions and related literature recommendations \cite{Guvenc18,Taha19}.
In addition to effective detection, classification and tracking methods, the existing literature also points out sensor fusion as a critical open area to achieve more accuracy and robustness compared to a single sensor.
Despite this, research in sensor fusion for drone detection is scarce \cite{Samaras_2019,Guvenc18,Diamantidou19,Shi18}.
Our work also encompasses collecting and annotating a public dataset to accomplish training and evaluation of the system.
Another fundamental challenge is the lack of public reference databases that serve as a benchmark for researchers \cite{Taha19}.
Three different consumer-grade drones are included in the dataset together with birds, airplanes and helicopters, in which constitutes the published dataset with the biggest number of target classes (drone, bird, airplane and helicopter, in comparison to others which only contain three, two or one of these classes only).
To achieve effective detection of drones, in building the classes, we have considered including other flying objects that are likely to be mistaken for a drone \cite{Saqib17,Aker17}.
Another missing piece in previous studies that we address here is the system's classification performance as a function of the distance to the target, with annotations of the database including such information as well.

A preliminary version of this article appeared at a conference~\cite{drone20icpr}.
In the present contribution, we substantially increase the amount of reported results, e.g. the previous paper only reported the precision, recall and F1-score of the individual sensors provided in Tables~\ref{tab:results-IRcam}, \ref{tab:results-Vcam} and \ref{tab:results-audio}, as well as the fusion results of Figure~\ref{fig:fusion-results1}.
Here, we extensively analyze the effect of internal parameters of the different detectors on their performance for the various sensors.
We also report results with a radar module and provide comments about the fish-eye camera motion detector, all of them missing in the previous publication.
Additional results on the fusion of sensors are also provided, including an Appendix with complementary observations and visual examples.
New detailed information about the system architecture, hardware and software employed is also provided, including details about implementation and design choices not included in the previous publication.
The related work is also described in more detail.

The rest of the paper is organized as follows.
Section~\ref{sect:soa} describes related work.
Section ~\ref{sect:materialsmethods} extensively describes the proposed system, including the architecture, the hardware components, the involved software, the Graphical User Interface, and the dataset. 
The experimental results are presented and discussed in Section~\ref{sect:results}. 
Finally, the conclusions are given in Section~\ref{sect:conclusions}.

\section{Related Work}
\label{sect:soa}


Fusing data from multiple sensors allows for more accurate results than a single sensor, while compensating for their individual weaknesses \cite{Samaras_2019}.
The sensors used for drone detection include:
$i$) radar (on several different frequency bands, both active and passive), $ii$) cameras in the visible spectrum, $iii$) cameras detecting thermal infrared emission (IR), $iv$) microphones to detect acoustic vibrations, i.e. sound, $v$) sensors to detect radio frequency signals to and from the drone and the controller (RF), and $vi$) scanning lasers (Lidar). As mentioned in \cite{Samaras_2019} and explored further in \cite{Boddhu13}, even humans are employed for the task. It has also been successfully demonstrated that animals can be trained for this role \cite{[GuardFromAboveBV]}.
Systems for drone detection utilizing one or more of the sensors mentioned above may also be combined with some effector to try to bring the flying drone down or take control of it in some other way. An effector component, however, is not part of this work.

An introduction to the subject and a comparison of drone detection and tracking techniques is given in the 2018 paper \cite{Guvenc18}. It highlights as open research the use of fusion techniques to exploit data from multiple sensors, and the development of effective machine learning techniques for detection, classification and tracking in various scenarios of interest. 
%
%
The paper also briefly provides an overview of ways to interdict unauthorized drones.

A comprehensive review of about 50 references is given in the 2019 paper \cite{Taha19}, which comments the different machine learning techniques based on the type of sensor, including its limitations.
The lack of public reference datasets is identified as an essential issue. Furthermore, no study analyzes the classification performance in relation to the distance to the drone. 
In addition, the sensing device, drone type, detection range, or dataset used is usually not specified, all key aspects to make works reproducible. 
%
%
%
%
%
%
%

Also from 2019, the paper \cite{Samaras_2019} has an exhaustive 178 references, not specific to drone detection and classification but also regarding the general techniques employed. 
%
%
It 
emphasizes as well the lack of works that use 
thermal cameras, despite the successful use of such sensors together with deep learning-based methods for general object detection. 
%

\subsection{Sensors Detecting Thermal Infrared Emission}

The work \cite{ANDRASI2017183}, from 2017, does not utilize machine learning, but 
a human looking at the output video stream.
The sensor is a low-cost FLIR Lepton 80$\times$60 pixels thermal camera. Connected to a Raspberry Pi, the authors are able to detect three different drone types up to 
100m. One conclusion of the paper is that the drone battery, and not the motors (as one may presume), is the most significant source of heat radiation.
With the background from this paper and the ones above, the present work will try to extend these findings using a higher resolution sensor (FLIR Boson with 320$\times$256 pixels) in combination with machine learning methods. 
The IR camera will also be combined with at least one additional sensor.

Thermal cameras combined with deep-learning detection and tracking are explored in the 2019 paper \cite{wang_chen_choi_kuo_2019}. The IR videos are of 1920$\times$1080, but the sensor is not specified.
Detection is done with a Faster-RCNN. Given the difficulties of acquiring enough data, it uses a modified Cycle-GAN (General Adversarial Network) to produce synthetic thermal training data.
Via precision-recall curves, the thermal detector is shown to achieve better performance than a visible sensor used for comparison.
Sadly, no details about the sensor-to-target distance are given.
The database is said to be public as ``USC drone detection and tracking dataset'', but without a link. The dataset is also mentioned in \cite{Wu18}, but the link in that paper is not working.
Compared to \cite{wang_chen_choi_kuo_2019}, the present paper uses three different drone types instead of one. We also expand the number of target classes to four and, additionally, we explore the detection performance as a function of sensor-to-target distance.

A thermal infrared camera is also used as one of the sensors in \cite{Diamantidou19}, but 
the paper fails to specify the type, field of view or even the resolution of the sensors used, so even if there are useful tables of the results, any comparison is unfortunately hard to achieve.

\subsection{Sensors in the Visible Range}

A widespread method to detect drones 
is to combine a video camera with a detector based on a convolutional neural network (CNN).
The paper \cite{Park17}, from 2017, studies six different CNN models, providing metrics for training time, speed performance (frames per second) and precision-recall curves. The comparison shows that considering the speed and accuracy trade-off, YOLOv2 \cite{Redmon17} seems to be the most appropriate detection model.

As it turns out from other studies on drone detection in the visible range, the YOLOv2 architecture is prevalent \cite{Wu18}, \cite{Liu_2018}, \cite{Saqib17} \cite{Aker17}. A lightweight version of the more recent YOLOv3 is utilized in \cite{Unlu19}.
The success of the YOLOv2 motivates its use in the present paper. This choice will also enable comparison to the papers mentioned above. 

The use of pan/tilt platforms to steer cameras in the direction of suspicious objects has also led to the use of wide-angle sensors. 
%
%
In \cite{Unlu19}, a static camera with 110$^{\circ}$ field of view (FoV) is used together with a YOLOv3 detector to align a rotating narrow-field camera. To find objects of interest with the wide-angle camera, the paper \cite{Unlu19} employs a Gaussian Mixture Model (GMM) foreground detector \cite{Stauffer99}, a strategy also followed in the present paper. However, our wide-angle sensor has an even wider FoV (180$^{\circ}$). As pointed out in \cite{Unlu19}, this setup is prone to produce false alarms in some situations, but as described later, 
this can be mitigated by tuning the associated detection and tracking algorithms.

Of all papers found using versions of the YOLO architecture for detection, \cite{Liu_2018} has the most output target classes with three (drone, airplane and helicopter), followed by \cite{Aker17} with two classes (drone and bird). However, none of the YOLO papers reports the detection accuracy as a function of the sensor-to-target distance.

\subsection{Acoustic Sensors}

Numerous papers have also explored the use of acoustic sensors. Some like \cite{Kim17}, \cite{Siriphun18}, \cite{Park15} utilize the Fast Fourier Transform (FFT) to extract features from the audio signals.
However, the Mel Frequency Cepstrum Coefficients (MFCC) seems to be the most popular technique, as used in \cite{Anwar19}, \cite{Liu17}, \cite{Jeon17}, \cite{Bernardini17}. The MFCC consists of a non-linear mapping of the original frequency according to the auditory mechanism of the human ear, and it is the most commonly used audio feature in current drone recognition tasks \cite{Liu_2018}.

When comparing 
classification models, 
the authors of \cite{Jeon17} conclude that Long Short-Term Memory (LSTM) \cite{Hochreiter97} achieves the best performance and F1-score. In that paper, the classification is binary (drone or background). The present work expands the range of output classes of \cite{Jeon17} by adding a helicopter class.
Also, the maximum acoustic detection range in the reviewed papers is 290 m, using a 120-element microphone array and a DJI Phantom 2 drone \cite{Busset15}. It is worth noting that the DJI Flame Wheel F450, one of the drones used in this work, is detected at a distance of 160 m by the microphone array.

\subsection{Radar}

Since radar is the most common technology to detect flying objects, it is not far-fetched to apply it to drones. 
However, a system designed to detect aircrafts often has features to reduce unwanted echoes from small, slow and low-flying objects, which is precisely what characterises UAVs. 
%
%
The small Radar Cross Sections (RCS) of medium-sized consumer drones are described in \cite{Patel18}, and from \cite{Herschfelt17} we have that the RCS of the DJI Flame Wheel F450 is -17 dBsm (0.02 m$^2$). The paper \cite{Gong19} points out that flying birds have similar RCS, which can lead to false targets. 
The F450 drone is also used in \cite{Fuhrmann17}, where the micro-doppler characteristics of drones are investigated. These are typically echoes from the moving blades of the propellers, and they can be detected on top of the bulk motion doppler signal of the drone. Since the propellers are generally made from plastic, the RCS of these parts are even smaller, and in \cite{Patel18}, it is stated that the echoes from the propellers are 20 to 25 dB weaker than the drone body itself. Nevertheless, papers like \cite{Bjorklund18}, \cite{Drozdowicz16} and \cite{Rahman18} accompany \cite{Fuhrmann17} in exploring the possibility of classifying drones using the micro-doppler signature.

\subsection{Other Drone Detection Techniques}

Very few drones are autonomous in the flight phase. Generally, they are controlled by ground equipment, and often send information on some radio frequency (RF), which can be used to detect them as well. The three drones used here are all controlled in real-time. The information sent out range from just simple telemetry such as battery level (DJI Flame wheel F450), a live video stream (Hubsan H107D+), to both a video stream and extensive position and status information (DJI Phantom 4 Pro).
Utilizing the RF fingerprint is described in \cite{Birnbach17}, and in \cite{Shorten18}, a CNN is used with data from an antenna array so that the direction to the drone controller can be calculated within a few degrees. In \cite{Ezuma20}, signals from 15 different drone controllers are classified with an accuracy of 98.13\% using only three RF features (shape factor, kurtosis and variance) with a K-Nearest Neighbour (KNN) classifier.
The use of LiDAR (Light Detection And Ranging) and LADAR (LAser Detection And Ranging) has also been explored \cite{Kim_2018}, 
successfully detecting drones up to 2 km. 

\subsection{Sensor Fusion}

The paper \cite{Samaras_2019} mentioned data fusion from multiple sensors as a way to improve accuracy, since sensor combination helps to compensate for individual weaknesses. 
As mentioned above, the benefits of sensor fusion are also pointed out in \cite{Guvenc18}.
Considering that missed detection of an intruding drone will bring more security threats than false alarms, the authors of \cite{Shi18} conduct audio, video and RF detection in parallel, using a logical OR operation to fuse the results of the three detectors.
They also highlight that combining such heterogeneous data sources is an open research question since the simple combination of the results separately obtained by audio, video, and RF surveillance can induce significant information loss. In contrast, it would be of greater significance to develop reliable fusion techniques at the feature level as well. 
%
%
Without any further specifications of the sensors used besides that they are visible, thermal and 2D-radar, the work \cite{Diamantidou19} presents promising results from experiments using a multilayer perceptron (MLP) to perform sensor fusion in a drone detector/classifier with just one output class.
Just as in \cite{Samaras_2019}, the present paper also considers early and late sensor fusion and differentiates these two principles based on whether the sensor data is fused before or after the detection element.

\subsection{Drone Detection Datasets}

As mentioned, the work \cite{Taha19} points out the lack of publicly available datasets. This is also highlighted in \cite{Samaras_2019}, especially with thermal infrared cameras.
%
%
The latter paper also states the strong need for real-world UAV audio datasets. 

In the references of \cite{Samaras_2019}, there are two useful links to datasets for visible video detectors. One of these is \cite{Reiser20}, where 500 annotated drone images can be found. 
This is far away from the 203k images of our database (plus audio clips).
The other link leads to the dataset \cite{SafeShore} of the drone-vs-bird challenge held by the Horizon2020 SafeShore project consortium. However, the dataset is only available upon request and with restrictions to the usage and sharing of the data. The drone-vs- bird challenge is also mentioned in \cite{Saqib17}, \cite{Aker17} and by the winning team of the 2017 challenge \cite{Schumann17}.

The dataset used in \cite{Unlu19} is not available due to confidentiality. Since the French Ministry of Defence funded the work, one can presume that the dataset, in one way or another, is a property of the French Government or the French Armed Forces.

\section{Materials and Methods}
\label{sect:materialsmethods}

This section describes the proposed methodology and outlines the automatic drone detection system, first on a system level and then in deeper detail. We detail the hardware components, how they are connected to the main computational resource, and the involved software running in the drone detection system, including the graphical user interface.
The methods used for the composition of our dataset are also described, including a description of the dataset, its division per sensor type, target class and sensor-to-target distance.
Our descriptive approach is motivated by \cite{Taha19}, which highlights that most works on visual drone detection do not specify the acquisition device, drone type, detection range, or dataset, all being key details that allow replication and comparison with other works.

\begin{figure} [htb]
\centering
\includegraphics[width=0.7\textwidth]{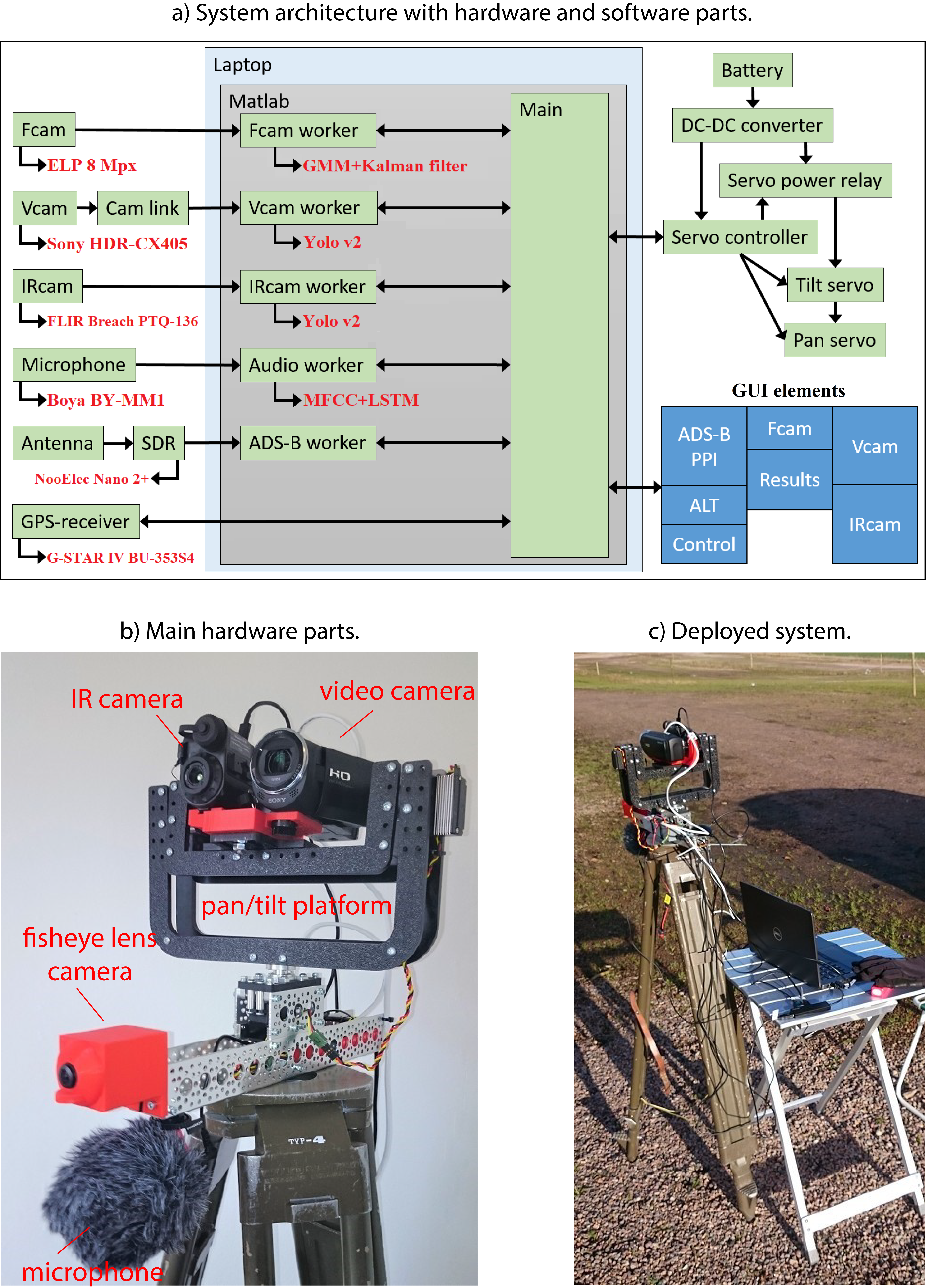}
\caption{System architecture, hardware and software parts. 
%
The deployed system is shown just north of the runway at Halmstad airport (IATA/ICAO code: HAD/ESMT). Pictures were originally appearing in \cite{drone20thesis} and published in \cite{drone20icpr}. Reprinted with permission.
\label{fig:system}}
\end{figure}

\subsection{Architecture of the System}

A drone detection system must be able to both cover a large volume of airspace and have sufficient resolution to distinguish small objects like drones and tell them apart from other types of objects.
Combining wide and narrow field of view (FoV) cameras is one way to accomplish this \cite{Unlu19}. Another way, shown in \cite{Liu_2018}, is to use an array of high-resolution cameras. 
Here, we follow the first approach since our solution has been designed with portability in mind (Figure~\ref{fig:system}c).
Since the present paper uses only one infrared sensor with a fixed FoV, there is no possibility of having neither a wide-angle infrared sensor nor an array of such sensors. The proposed way to achieve the desired volume coverage with the IR-sensor is to have it on a moving platform, as shown in Figure~\ref{fig:system}b. This platform can either have objects assigned to it or search by itself at moments when the sensors are not busy detecting and classifying objects.
The overall system architecture, detailing the hardware and software components employed, is shown in Figure~\ref{fig:system}a. The layout of the Graphical User Interface (GUI) elements is shown in the bottom right part (boxes in blue).

To be able to react to moving objects and also to have the ability to track those, the combined time constraints of the detection cycle and the control loop of the moving platform means that the system must work in close to real-time. Hence, all the detection and classification processes must be done efficiently and with as little delay as possible. The feedback loop of the moving platform must run at a sub-second speed.
In putting together such a system involving several sensors and mechanical parts, choosing the proper methods is critical. 
All these constraints, in turn, impose demands on the efficiency of the software as well.
Another aspect is that to detect the drones with high efficiency, the system must also recognize and track other flying objects that are likely to be mistaken for drones. For some of these drone-like objects, this is indigenous hard, e.g. birds. For others, it is technically possible since some of them announce their presence and location via radio, e.g. ADS-B, over which most aircrafts regularly transmit messages with varied content.
Combining the data from several sensors under the time constraints described above must be kept simple and streamlined too. This, together with the fact that very few papers have explored sensor fusion techniques, is the motivation to have a system where the inclusion and weights of the sensors can be altered at runtime to find a feasible setting.

\subsection{Hardware}

As primary electro-optical sensors, we use a thermal infrared camera (denoted as IRcam in Figure~\ref{fig:system}a) and a video camera (Vcam). 
Our system can keep track of cooperative aircrafts via ADS-B information that is made available with an antenna, which collects the aircraft’s position, velocity vectors and identification information broadcasted by aircrafts equipped with such a system.
We also include audio information through a microphone, which is employed to distinguish drones from other objects in the vicinity, such as helicopters.
All computations are made in a standard laptop, also used to present the results to the user via the designed GUI.

Since the primary cameras have a limited FoV, a fish-eye lens camera (Fcam) covering 180$^{\circ}$ horizontally and 90$^{\circ}$ vertically is also used. The fish-eye camera is used to detect moving objects in its FoV and then steer the IRcam and Vcam towards the detected objects using the pan/tilt platform. 
If the Fcam detects nothing, the platform can be set to move in two different search patterns to scan the sky.
As additional sensors, our system also includes a GPS receiver.
The hardware components are mounted on a standard surveyor tripod to provide stability to the system. This solution also facilitates the deployment of the system outdoors, as shown in Figure~\ref{fig:system}c. 
Due to the nature of the system, it must also quickly be transported to and from any deployment. Hence, 
the system can be disassembled into a few large parts and placed in a transport box.

\subsubsection{Thermal Infrared Camera (IRcam)}

We employ a FLIR Breach PTQ-136 using the Boson 320$\times$256 pixels detector (Y16 with 16-bit grey-scale). The FoV of the IRcam s 24$^{\circ}$ horizontally and 19$^{\circ}$ vertically.
%
%
It is worth noting that this sensor has a higher resolution than the FLIR Lepton sensor with 80$\times$60 pixels used in \cite{ANDRASI2017183}. 
In that paper, the authors detected three drone types up to a distance of 100m. However, it was done manually by a person looking at the live video stream. 
In contrast, the present paper employs an automatic detection solution.
The signal of the IRcam is sent to the laptop at 60 frames per second (FPS) using the USB-C port, which also powers the IRcam. 

\subsubsection{Video Camera (Vcam)}

To capture video in the visible range, we employ a Sony HDR-CX405 video camera.
The feed is taken from the HDMI port, which is captured with an Elgato Cam Link 4K frame grabber that provides a 1280$\times$720 video stream in YUY2-format (16 bits per pixel) at 50 FPS.
The FoV of the Vcam can be made wider or narrower using its adjustable zoom lens.
In this work, it is adjusted to have about the same FoV as the IRcam. 

\subsubsection{Fish-eye Lens Camera (Fcam)}

To counteract the limited FoV of the IRcam and Vcam, a fish-eye lens camera is used to monitor a wider area of the sky and then steer and focus these two towards the detected objects to ascertain if they are drones or something else. 
The fish-eye lens camera employed is an ELP 8 Megapixel with a FoV of 180$^{\circ}$ degrees, which provides a 30FPS 1024$\times$768 video stream in Mjpg-format at 30 FPS via USB.

\subsubsection{Microphone}

The microphone is used to distinguish drones from other objects emitting sounds, such as, for example, helicopters. Here, we use a Boya BY-MM1 mini cardioid directional microphone connected directly to the laptop.
Data is stored in .wav format, with a sampling frequency of 44100 Hz.

\subsubsection{ADS-B Receiver}

To track aircraft equipped with transponders, an ADS-B receiver is also used. This consists of an antenna and a NooElec Nano 2+ Software Defined Radio receiver (SDR). This is tuned to 1090 MHz so that the identification and positional data sent out as a part of the 1 Hz squitter message can be decoded and displayed. The Nano 2+ SDR receiver is connected to the laptop via USB.

\subsubsection{GPS receiver}

To correctly present the decoded ADS-B data, the system is equipped with a G-STAR IV BU-353S4 GPS receiver connected via USB. The receiver outputs messages following the National Marine Electronics Association (NMEA) standard.

\begin{figure} [htb]
\centering
\includegraphics[width=0.6\textwidth]{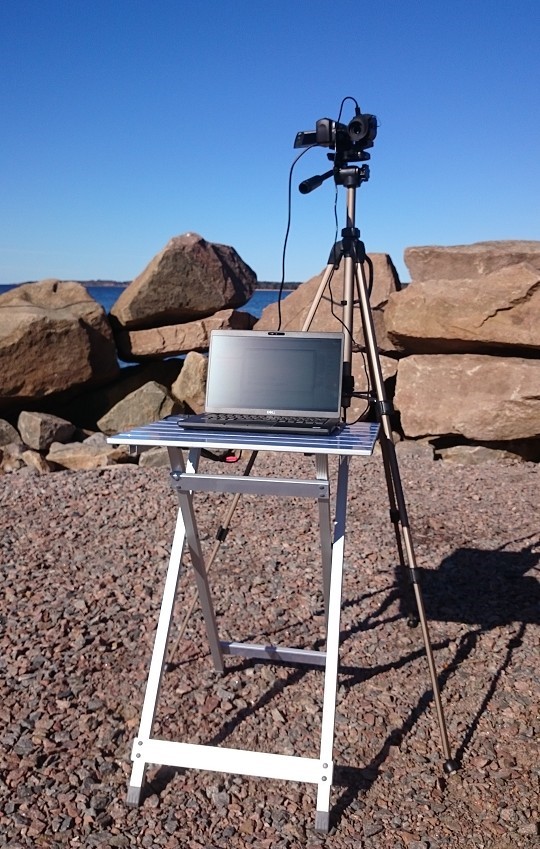}
\caption{The data collection setup using a lighter version of the system. Picture originally appearing in \cite{drone20thesis}. \label{fig:data_collection}}
\end{figure}

\subsubsection{Pan/tilt platform and servo controller}

The pan/tilt platform is a Servocity DDT-560H direct drive tilt platform together with the DDP-125 Pan assembly, also from Servocity. To achieve the pan/tilt motion, two Hitec HS-7955TG servos are used.
A Pololu Mini Maestro 12-Channel USB servo controller is included so that the respective position of the servos can be controlled from the laptop. Since the servos have shown a tendency to vibrate when holding the platform in specific directions, a third channel of the servo controller is also used to give the possibility to switch on and off the power to the servos using a small optoisolated relay board.

To supply the servos with the necessary voltage and power, both a net adapter and a DC-DC converter are available. The DC-DC solution is used when the system is deployed outdoors, and, for simplicity, it uses the same battery type as one of the available drones.
Some other parts from Actobotics are also used in mounting the system, and the following have been designed and 3D-printed: adapters for the IR, video and fish-eye lens cameras, a radar module mounting plate and a case for the servo controller and power relay boards.

A lighter version of the IRcam and Vcam mounting without the pan/tilt platform has also been prepared. This is used on a lightweight camera tripod when collecting the dataset, simplifying transportation and giving the possibility to set its direction manually. The data collection setup is shown in Figure~\ref{fig:data_collection}.

An unforeseen problem occurring when designing the system was actually of mechanical nature. Even though the system uses a pan/tilt platform with ball-bearings and very high-end titanium gear digital servos, the platform was observed to oscillate in some situations. This phenomenon was mitigated by carefully balancing the tilt platform and introducing some friction in the pivot point of the pan segment. It might also be the case that such problems could be overcome using a servo programmer. Changing the internal settings of the servos could also increase their maximum ranges from 90$^{\circ}$ to 180$^{\circ}$. This would extend the volume covered by the thermal infrared and video cameras so that all targets tracked by the fish-eye camera could be investigated, not just a portion of them, as now.

\subsubsection{Computer}

A Dell Latitude 5401 laptop handles the computational part of the system. It is equipped with an Intel i7-9850H CPU and an Nvidia MX150 GPU. The computer is connected to all the sensors mentioned above and the servo controller using the built-in ports and an additional USB hub, as shown in Figure~\ref{fig:system}c.

It is observed, regarding the sensors, that there is a lower limit of around 5 FPS, where the system becomes so slow that the ability to track flying objects is lost. All actions taken by the system must be well balanced, and just such a simple thing as plotting the ADS-B panel with a higher frequency than necessary can cause a drop in the FPS rate. Such problems can be overcome by using more than one computational resource.

\subsubsection{Radar module}

\begin{figure} [htb]
\centering
\includegraphics[width=0.6\textwidth]{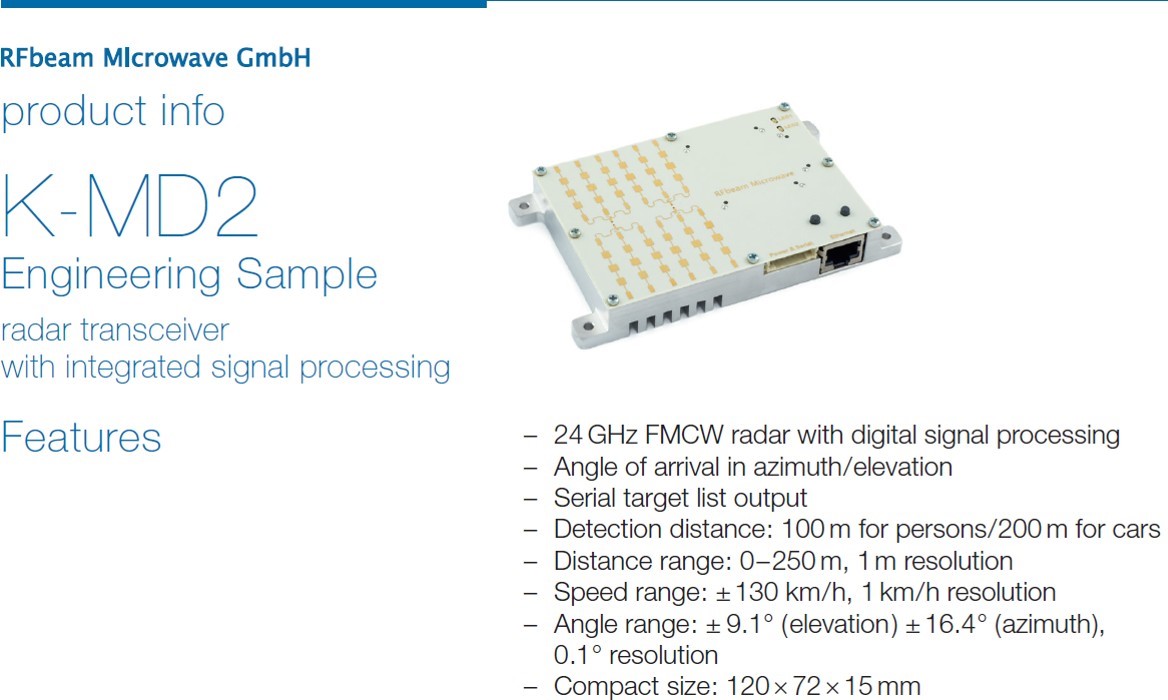}
\caption{The radar module K-MD2. Picture taken from\\ https://www.rfbeam.ch/files/products/21/downloads/Datasheet\_K-MD2.pdf \label{fig:radar}}
\end{figure}

As indicated earlier, our solution does not include a radar module in its final deployment. 
However, since one was available, it was included in our preliminary tests. 
It is a radar module from K-MD2, whose specifications are shown in Figure~\ref{fig:radar} \cite{RFbeam}.
Its exclusion was motivated by its short practical detection range.
Interestingly, the K-MD2 radar module is also used in another research project connected to drones \cite{Mostafa18}, not to detect drones, but instead mounted on board one as part of the navigation aids in GNSS2 denied environments.

\subsection{Software}

Matlab is the primary development environment in which the drone detection system has been developed.
The software for drone detection consists of the main script and five separate `workers', one per sensor, as shown in Figure~\ref{fig:system}a.
The main script and the workers can run asynchronously and in parallel, enabled by the Matlab parallel computing toolbox. 
This allows each detector to run independently of the others. This also allows the different sensors to run asynchronously, handling as many frames per second as possible without inter-sensor delays and waiting time.
The main script communicates with the workers using pollable data queues.
The \textit{Fcam worker} utilizes a foreground/background detector via GMM background subtraction \cite{Stauffer99,gmmmatlab} and a multi-object Kalman filter tracker \cite{kalmanmatlab}. After calculating the position of the best-tracked target (defined as the one with the longest track history), it sends the azimuth and elevation angles to the main script, which then controls the pan/tilt platform, so that the moving object can be analysed further by the IR and video cameras.
The \textit{IRcam} and \textit{Vcam} workers are similar in their basic structure, and both import and run a trained YOLOv2 detector, fine-tuned with annotated ground truth to work with data from each camera.
The information sent to the main script is the region of the image where the object has been detected, the class of the detected target found in the ground truth of the training dataset, the confidence, and the horizontal and vertical offsets in degrees from the centre point of the image. The latter is used to calculate servo commands and track the object.
The \textit{Audio} worker sends information about the class and confidence to the main script. 
It uses a classification function built on an LSTM architecture, which is applied to MFCC features extracted from audio signals captured with the microphone. 
Unlike the others, the \textit{ADS-B} worker has two output queues, one consisting of current tracks and the other of the history tracks. 
The ``current'' queue contains the current positions and additional information (Id, position, altitude, calculated distance from the system, azimuth and elevation in degrees as calculated relative to the system, time since the message was received and the target category).
The ``history'' tracks queue is just a list of the old positions and altitudes.
This partition in two queues saves computational resources by reducing the amount of data from the worker drastically, so that only the information needed/used by the main process is sent. 
It also makes easier to control the data flow, since the length of the history track queue is easily set if it is separated.
%
%
All of the above workers also send a confirmation of the command from the main script to run the detector/classifier or to be idle. The number of frames per second currently processed is also sent to the main script.

Table~\ref{tab:output-classes}
shows the different classes that each worker can provide to the main script.
Note that not all sensors can output all the target classes. 
The audio worker has an additional ``background'' class, and the ADS-B will output a ``no data'' class if the vehicle category field of the received message is empty (since it is not a mandatory field of ADS-B messages).

\begin{table}[htb]
\caption{Output classes of the sensors and their corresponding class colours. Table originally appearing in \cite{drone20thesis}. \label{tab:output-classes}}

\begin{center}
\begin{tabular}{c}

\includegraphics[width=0.85\textwidth]{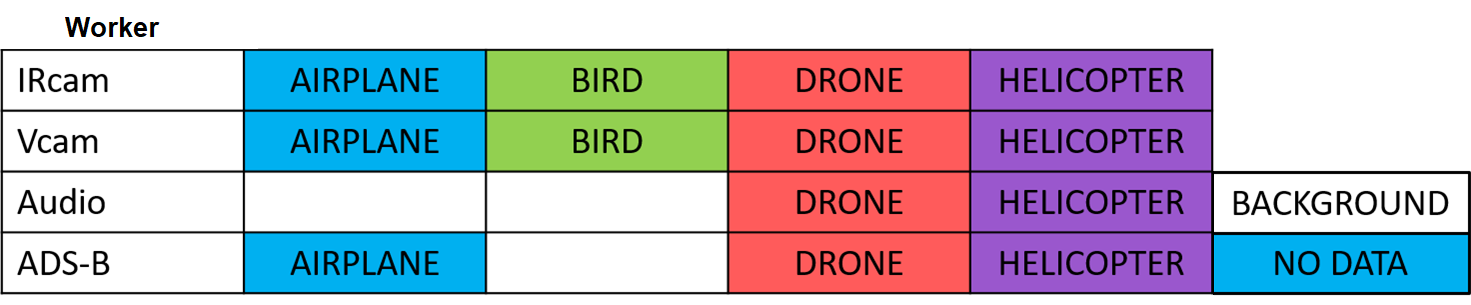}  \\ 

\end{tabular}

\end{center}

\end{table}
\normalsize

%
%
%

\subsubsection{Main Script}

This is the core of the system. Besides starting the five workers (threads) and setting up the queues to communicate with these, it also sends commands to and reads data from the servo controller and the GPS receiver.
After the start-up sequence, the script goes into a loop that runs until the program is stopped by the user via the graphical user interface (GUI).

Updating the GUI and reading user inputs are the most frequent tasks on every loop iteration. The main script interacts with the workers and the servo controller at regular intervals. Servo positions are read, and queues are polled ten times a second. The system results, i.e. the system output label and confidence, are also calculated using the most recent results from the workers. Furthermore, at a rate of 5 Hz, new commands are sent to the servo controller for execution. Every two seconds the ADS-B plot is updated. Different intervals for various tasks make the script more efficient since, for example, an aircraft sends out its position via ADS-B every second. Hence, updating the ADS-B plots too often would only be a waste of computational resources. 

\subsubsection{IRcam worker}

Raw images are processed with Matlab function \texttt{imflatfield}\footnote{2-D image flat-field correction using Gaussian smoothing with a standard deviation of $\sigma$ to approximate the shading component of the input image. It cancels the artefacts caused by variations in the pixel-to-pixel sensitivity of the detector and by distortions in the optical path. As a result, the corrected image has more uniform brightness.} for 2-D image flat-field correction (with $\sigma$=30) followed by \texttt{imadjust}\footnote{Increases the contrast of the image by mapping intensity values so that 1\% of the data is saturated at low and high intensities} to increase image contrast.
Flat-field correction uses Gaussian smoothing to approximate the shading component of the input image.
Next, the input image is processed by the YOLOv2-detector, with a given detection threshold and the execution environment set to GPU. The output from the detector consists of an array of class labels, confidence scores and bounding boxes for all objects detected and classified. The detector output may be no data at all, or just as well, data for several detected objects. In this implementation, only the detection with the highest confidence score is sent to the main script.
Images from the IR camera are 320$\times$256 pixels.
To present the result in the GUI at the same size as the Vcam output, the image is resized to 640$\times$512.
Then, the bounding box, class label, and confidence score are inserted into the image. To clearly indicate the detected class, the inserted annotation uses the same colour scheme as in Table~\ref{tab:output-classes}.
The current state of the detector and its performance (frames per second) is also inserted in the top left corner of the image. Such information is also sent to the main script with the detection results.

\begin{figure} [htb]
\centering
\includegraphics[width=0.7\textwidth]{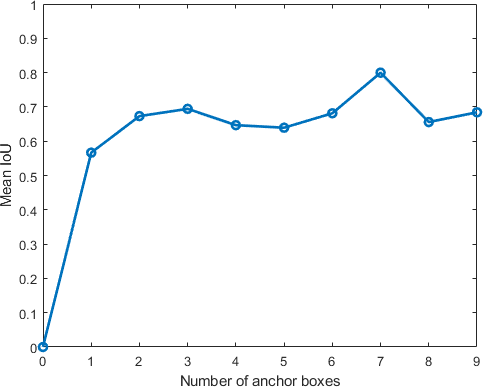}
\caption{Plot used to assess the number of anchor boxes to be implemented in the IRcam worker YOLOv2 detector. Picture originally appearing in \cite{drone20thesis}.}
\label{fig:IRcam-anchors-yolo}
\end{figure}

The YOLOv2 detector is formed by modifying a pretrained MobileNetv2 following \cite{yolo2matlab} so that the first 12 layers out of 53 are used for feature extraction. 
The input layer is changed to 256$\times$256$\times$3. 
Six detection layers and three final layers are also added to the network. Besides setting the number of output classes of the final layers, the anchor boxes used are also specified.
To estimate the numbers of anchor boxes to use and the size of these, the training data is processed using the \texttt{estimateAnchorBoxes} Matlab function. This function uses a $k$-means clustering algorithm to find suitable anchor boxes sizes given the number of boxes to be used, returning as well the mean intersection-over-union (IoU) value of the anchor boxes in each cluster. We test the number of anchor boxes from one to nine to provide a plot over the IoU as a function of the number of boxes, as shown in Figure~\ref{fig:IRcam-anchors-yolo}.

When choosing the number of anchor boxes to use, the trade-off to consider is that a high IoU ensures that the anchor boxes overlap well with the bounding boxes of the training data, but, on the other hand, using more anchor boxes will also increase the computational cost and may lead to over-fitting. After assessing the plot, the number of anchor boxes is chosen to be three, and the sizes of these (with the scaling factor of 0.8 in width to match the downsize of the input layer from 320 to 256 pixels) are taken from the output of the \texttt{estimateAnchorBoxes} function.

The detector is trained with \texttt{trainYOLOv2ObjectDetector}\footnote{Returns an object detector trained using YOLO v2 with the specified architecture, in our case a pretrained MobileNetv2 \cite{yolo2matlab} with the modifications mentioned in the main text.} using data picked from the available dataset (Section~\ref{sect:results}).
%
The detector is trained for five epochs using the stochastic gradient descent with momentum (SGDM) optimizer and an initial learning rate of 0.001.
Using a computer with an Nvidia GeForce RTX2070 8GB GPU, the time for one epoch is 39 min. 
The training function includes pre-processing augmentation consisting of horizontal flipping (50\% probability), scaling (zooming) by a factor randomly picked from a continuous uniform distribution in the range [1, 1.1], and random colour jittering for brightness, hue, saturation, and contrast.

\subsubsection{Vcam worker}

To be able to compare their results, the methods and settings of the Vcam worker are very similar to the IRcam worker above, with some exceptions. 
Images are of 1280$\times$720 pixels, which are then resized to 640$\times$512 without further pre-processing. 
Given the bigger image size, the input layer of the YOLOv2 detector here has a size of 416$\times$416$\times$3. 
%
%
Due to the increased image size, the training time is also extended compared to the IR case. When using a computer with an Nvidia GeForce RTX2070 8GB GPU, the time for one epoch is 2 h 25 min. 
%
The detector is trained for five epochs, just like the IRcam detector.

\subsubsection{Fcam worker}

Initially, the fish-eye lens camera was mounted upwards, but this caused the image distortion to be significant in the area just above the horizon where the interesting targets usually appear. After turning the camera so that it faces forward (as seen in Figure \ref{fig:system}b), the motion detector is less affected by image distortion.

The images from the camera are of 1280$\times$768 pixels, but the lower half (showing the area below the sky horizon) is cropped so that 1024$\times$384 pixels remain to be processed. 
Images are then analysed using the Matlab \texttt{ForegroundDetector} function \cite{gmmmatlab}, which compares an input video frame to a background model to determine whether individual pixels are part of the background or the foreground. 
The function uses a background subtraction algorithm based on Gaussian Mixture Models (GMM) \cite{Stauffer99}, producing a binary mask with pixels of foreground objects set to one.
The mask is next processed with the \texttt{imopen} function, which performs a morphological opening (erosion followed by dilation) to eliminate noise. The structural element is set to 3$\times$3, so that very small objects are deleted.
Then, the \texttt{BlobAnalysis} function is applied, which outputs the centroids and bounding boxes of all objects in the binary mask provided by the \texttt{ForegroundDetector} function. 
All centroids and bounding boxes are sent finally to a multi-object tracker based on Kalman filters \cite{kalmanmatlab}, created with the \texttt{configureKalmanFilter} function, which tracks objects across multiple frames. 
A Kalman filter is used to predict the centroid and bounding box of each track in a new frame based on their previous motion history. Then, the predicted tracks are assigned to the detections given by the foreground/background detector by minimizing a cost function that takes into account the Euclidean distance between predictions and detections.
The object with the longest track history is picked and retained as the best. 

\begin{figure} [htb]
\centering
\includegraphics[width=0.85\textwidth]{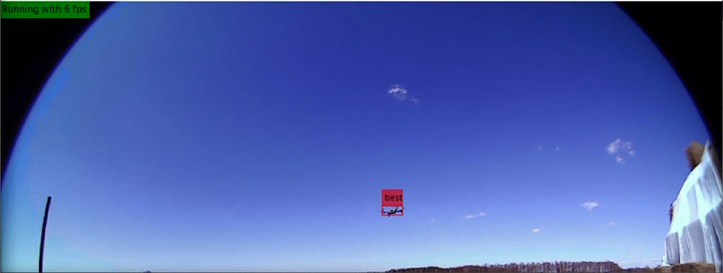}
\caption{Fish-eye camera image with a detected track in red. Picture originally appearing in \cite{drone20thesis}.}
\label{fig:Fcam_red}
\end{figure}

In the Fcam presentation window, all tracks (both updated and predicted) are visualised, and the track considered to be the best is marked with red (Figure~\ref{fig:Fcam_red}).
The output from the Fcam worker is the FPS status, together with the elevation and azimuth angles of the best track, if such track exists at the moment. 
Of all the workers, the Fcam is the one with the most tuning parameters, as will be seen in Section~\ref{sect:results}. This involves choosing and tuning the image processing operations, foreground detector and blob analysis settings, and the multi-object Kalman filter tracker parameters.

\subsubsection{Audio worker}

The audio worker collects acoustic data in a one-second-long buffer (44100 samples), set to be updated 20 times per second. 
To classify the sound source in the buffer, it is first processed with the Matlab \texttt{mfcc} function, which extracts MFCC features. 
We employ the default parameters (a Hamming window with a length of 3\%  and an overlap of 2\% of the sampling frequency, and the number of coefficients per window equal to 13). 
Based on empirical trails, the parameter \texttt{LogEnergy} is set to \texttt{Ignore}, meaning that the log-energy is not calculated. Then the extracted features are sent to the classifier.
The extracted features are then sent to an LSTM classifier consisting of an input layer, two bidirectional LSTM layers with a dropout layer in-between, a fully connected layer, a softmax layer and a classification layer. The classifier builds on \cite{[LSTM]} but increasing the number of classes from two to three and an additional dropout layer between the bidirectional LSTM layers.

\begin{figure} [htb]
\centering
\includegraphics[width=0.85\textwidth]{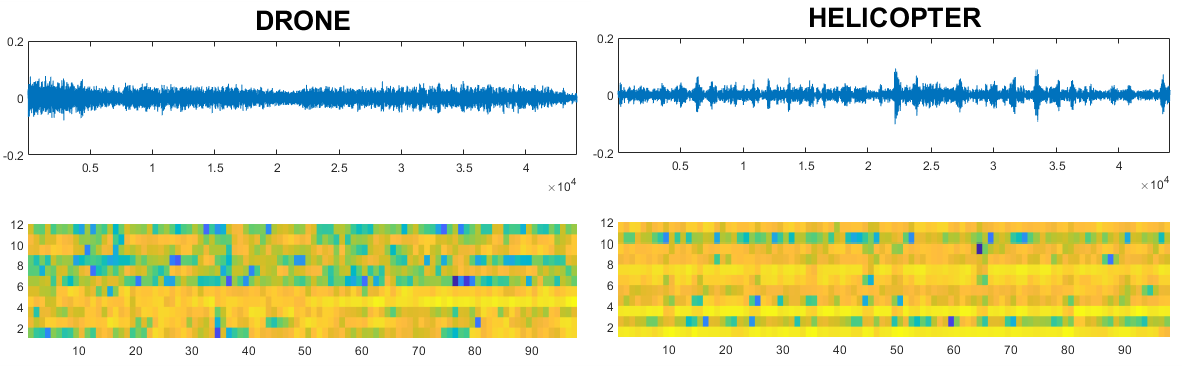}
\caption{Two examples of the audio worker plots with the audio  amplitudes and the extracted MFCC features. Pictures originally appearing in \cite{drone20thesis}.}
\label{fig:audio-features}
\end{figure}

%
The classifier is trained from scratch for 120 epochs, after which it shows signs of over-fitting. 
%
%
It includes a class of general background sounds (Table~\ref{tab:output-classes}) recorded outdoors in the typical deployment environment of the system. Also, it has some clips of the sounds from the servos moving the pan/tilt platform. Like the other workers, the output is a class label with a confidence score.
A graphical presentation of the audio input and extracted features is also available, as shown in Figure~\ref{fig:audio-features}, but this is not included in the final GUI layout.

\subsubsection{ADS-B worker}

We implement the ADS-B decoding in Matlab using functions from the Communications toolbox ~\cite{adsbmatlab}.
All vehicle categories that can be seen as subclasses to the airplane target label are clustered together. These are all the classes ``Light'', ``Medium'', ``Heavy'', ``HighVortex'', ``VeryHeavy'' and ``HighPerformanceHighSpeed''. The class ``Rotorcraft'' is translated into helicopter. Interestingly, there is also a ``UAV'' category label. This is also implemented in the ADS-B worker, translated into drone.

\begin{figure} [htb]
\centering
\includegraphics[width=0.95\textwidth]{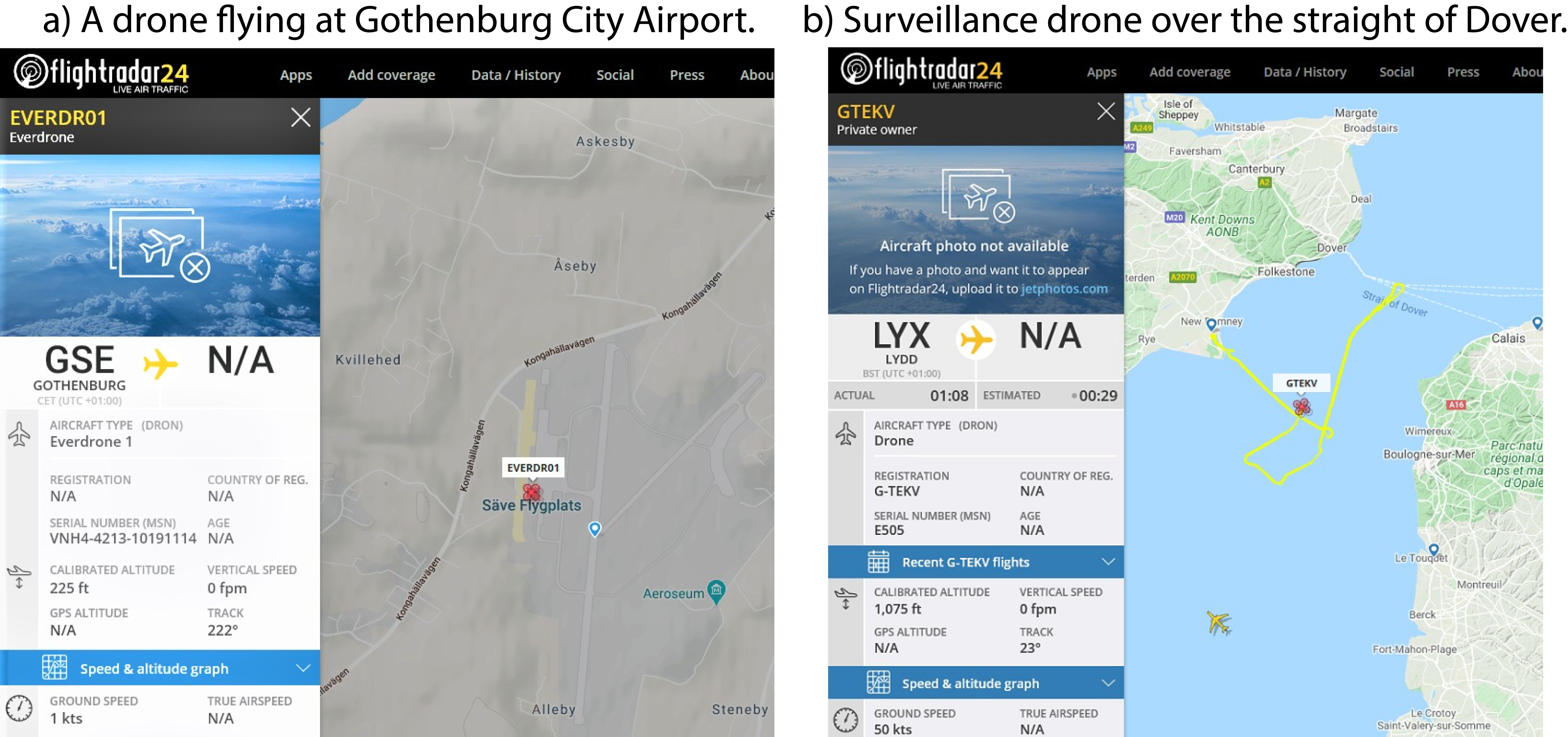}
\caption{Example of drones sending ADS-B information.
Images from www.flightradar24.com}
\label{fig:flightradar}
\end{figure}

One might wonder if there are any such aircraft sending out that belong to the UAV vehicle category. Examples are found by looking at the Flightradar24 service (Figure~\ref{fig:flightradar}). Here we can find one such drone flying at Gothenburg City Airport, one of the locations used when collecting the dataset. The drone is operated by the company Everdrone AB, involved in the automated external defibrillators delivery trails of \cite{[Sanfridsson19]}.
Another example is the UK Coastguard/Border Force surveillance drone that is regularly flying at night over the straight of Dover since December of 2019. This is naturally a large drone with a wingspan of 7.3 m.

As mentioned above, not all aircrafts will send out their vehicle category as part of the ADS-B squitter message. 
However, in our experience, about half of the aircrafts send out their category. This justifies its inclusion in this work, as one of our principles is to detect and keep track of other flying objects that are likely to be mistaken by a drone.
In the output message of the ADS-B worker, the confidence of the classification is set to 1 if the vehicle category message has been received. If not, the label is set to airplane (the most common aircraft type) with the confidence to 0.75 so that there is a possibility for any of the other sensors to influence the final classification.

\subsection{Graphical User Interface (GUI)}
\label{sect:GUI}

\begin{figure} [htb]
\centering
\includegraphics[width=0.95\textwidth]{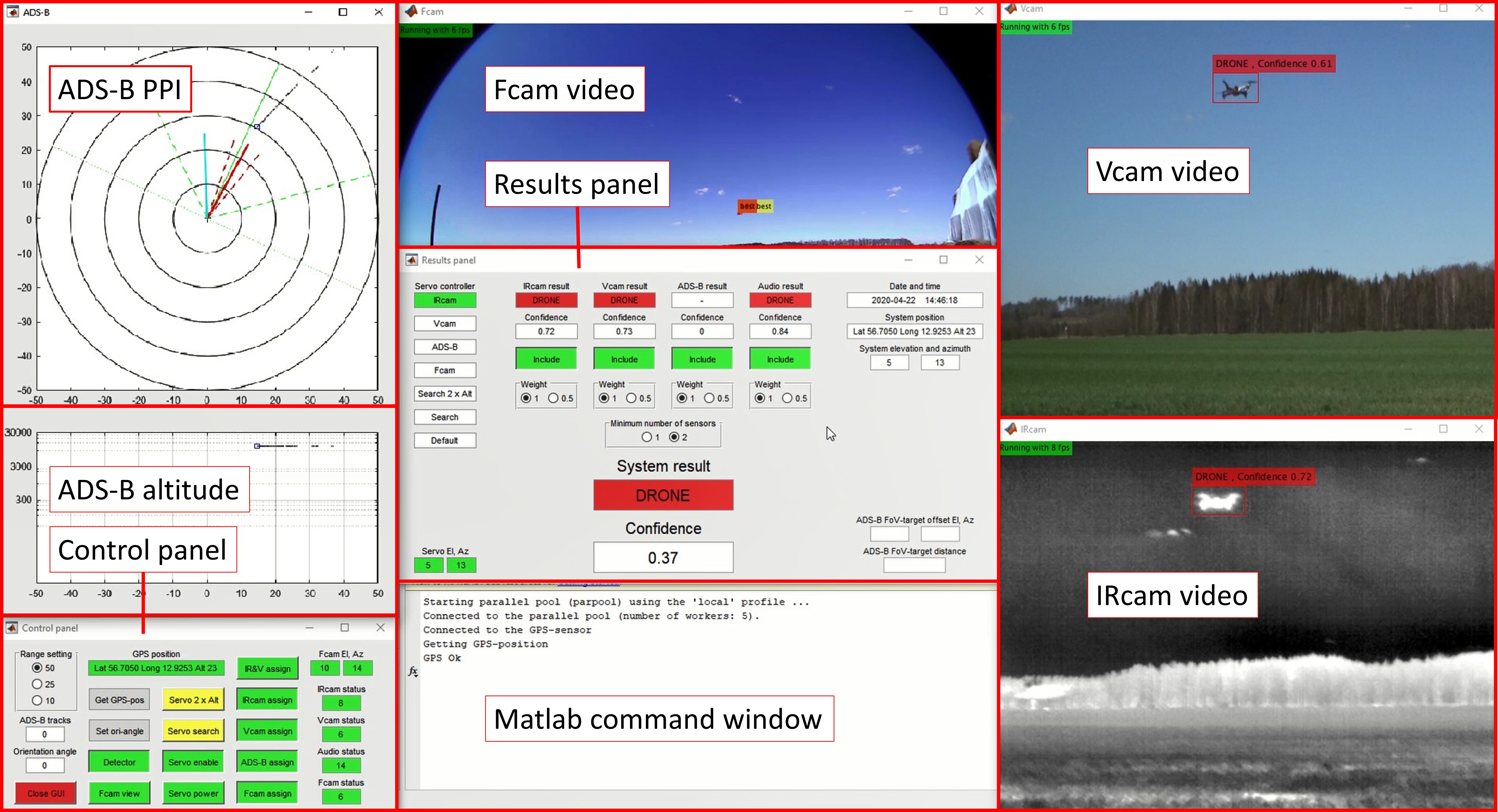}
\caption{Graphical User Interface of the system with labels of the different elements. 
Picture originally appearing in \cite{drone20thesis} and published in \cite{drone20icpr}. Reprinted with permission.}
\label{fig:system_GUI}
\end{figure}

The Graphical User Interface of the system is shown in Figure~\ref{fig:system_GUI}.
The GUI is part of the main script but is described separately. 
It shows the output of the different workers, including video streams of the cameras, providing various options to control the system configuration.
The Matlab command window is made visible in the bottom centre so that messages (e.g. exceptions) can be monitored during the development and use of the system.

The ADS-B presentation area (left) consists of a PPI-type (Plan Position Indicator) display and an altitude display. Besides presenting the ADS-B targets, the PPI display also shows system information. The solid green line is the main direction of the system relative to the north. The other green lines present the field of motion of the pan/tilt platform (dashed) and the field of view of the fish-eye lens camera (dotted).
The actual direction of the pan/tilt platform is presented with a solid red line, and the field of view of the thermal infrared and video cameras are represented using dashed red lines. If the fish-eye lens camera worker tracks any object, its direction is indicated by a solid cyan line.
ADS-B targets are presented using the class label colours of Table~\ref{tab:output-classes}, together with the track history plots. The altitude information is presented in a logarithmic plot to make the lower altitude portion more prominent.

\begin{figure} [htb]
\centering
\includegraphics[width=0.95\textwidth]{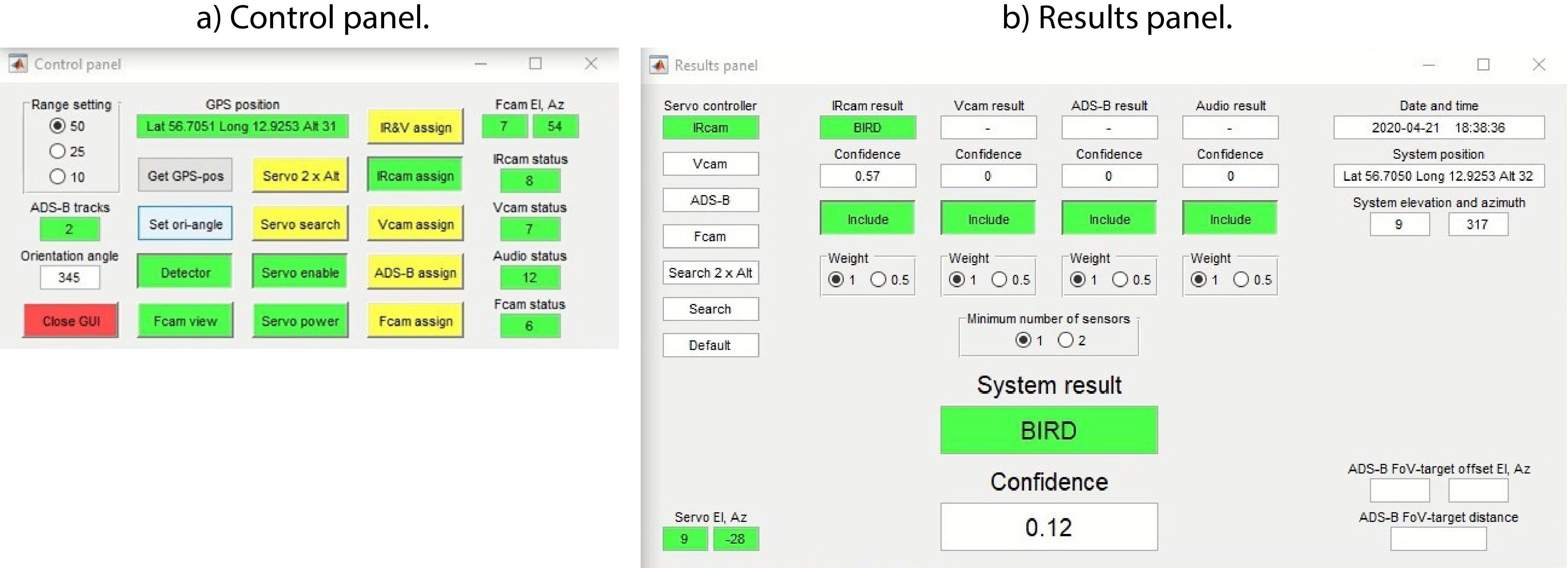}
\caption{Detail of the control panel and results panel of the Graphical User Interface. In the control panel, note a bird detected and tracked by the IRcam worker. Pictures originally appearing in \cite{drone20thesis}.}
\label{fig:gui-fig18-fig19}
\end{figure}

The area directly below the ADS-B presentation is the control panel (seen in better detail in Figure~\ref{fig:gui-fig18-fig19}a).
Starting from the top left corner, we have radio buttons for the range settings of the ADS-B PPI and altitude presentations. Next is the number of ADS-B targets currently received, and below that, the set orientation angle relative to the north. The ``Close GUI'' button is used to shut down the main script and the workers.
The GPS-position presentation changes colour to green when the GPS receiver has received a correct position after pressing the ``Get GPS-pos'' button. Pressing the ``Set ori-angle'' button opens an input dialogue box so that the orientation angle of the system can be entered. Below that two buttons for switching the detectors between running and idle mode and a choice to display the raw Fcam image or the moving object mask only (not shown).
The servo settings can be controlled with the buttons of the middle column.
To aid the Fcam in finding targets, the pan/tilt can be set to move in two different search patterns. One where the search is done from side to side using a static elevation of 10$^{\circ}$, so that the area from the horizon up to 20$^{\circ}$ is covered, and another one where the search is done with two elevation angles to increase the coverage.
The pan/tilt platform can also be controlled by the elevation and azimuth angles from one of the workers. This is set by the ``assign'' buttons of the fourth column, placed in priority order from top to down. The ``IR\&V assign'' setting means that a target has to be detected by both the IRcam and Vcam workers, and if so, the pan/tilt platform is controlled by the angular values from the IRcam worker.
The rightmost column of the control panel shows the status information regarding the performance in FPS of the workers and the elevation and azimuth angles of the Fcam worker target (if such a target exists). The status displays are red if the worker is not connected, yellow if the detector is idle, and green if it is running.

The results panel (seen in better detail in Figure~\ref{fig:gui-fig18-fig19}b) features settings for the sensor fusion and presents the workers and system results.
The first column (servo controller) indicates the source of information currently controlling the servos of the pan/tilt platform.
In the bottom left corner, the angles of the servos are presented. 
The settings for the sensor fusion and the detection results presentation are found in the middle of the panel. 
The information in the right part of the panel is the current time and the position of the system. The system elevation and azimuth relative to the north are also presented here. Note the difference in azimuth angle compared to the bottom left corner where the system internal angle of the pan/tilt platform is presented.
The last part of the results panel (bottom right corner) presents offset angles for the ADS-B target if one is in the field of view of the thermal infrared and video cameras. These values are used to detect systematic errors in the orientation of the system. The sloping distance to the ADS-B target is also presented.

\subsection{Database}
\label{sect:database}

\begin{figure} [htb]
\centering
\includegraphics[width=0.85\textwidth]{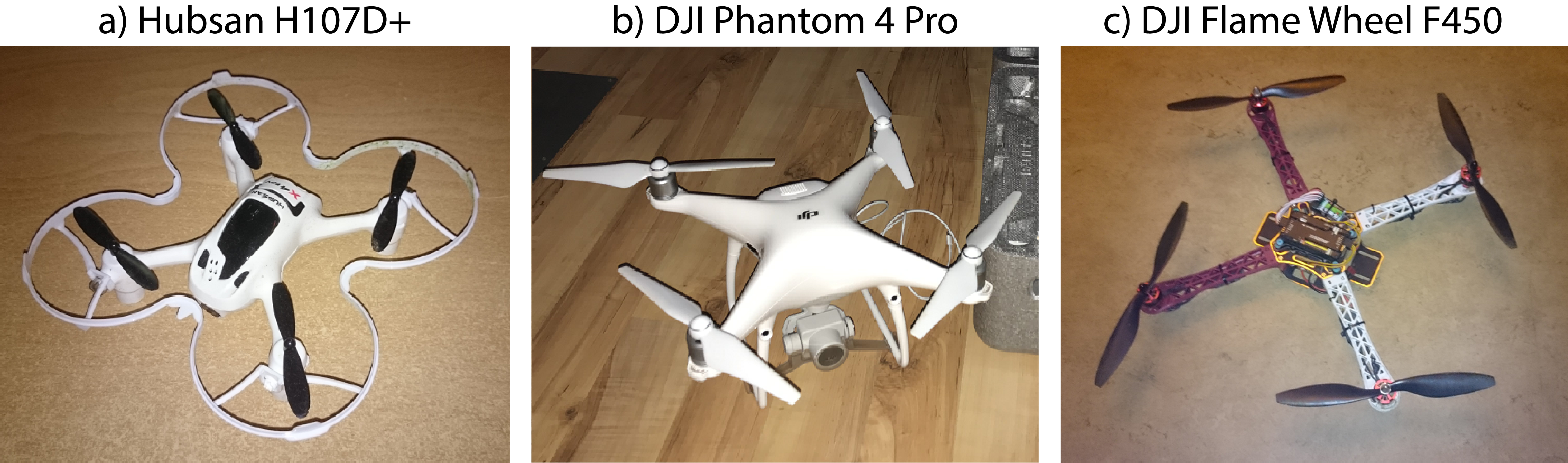}
\caption{The three drones of our dataset. 
Pictures were originally appearing in \cite{drone20thesis} and published in \cite{drone20icpr} and \cite{DiBdataset}. Reprinted with permission.}
\label{fig:db_drones}
\end{figure}

A dataset has also been collected to accomplish the necessary training of the detectors and classifiers. The annotation has been done so that others can inspect, edit, and use the dataset. The fact that the datasets for the thermal infrared and the visible video sensors are collected under the same conditions and using the same targets ensures that a comparison between the two sensor types is well-founded.
%
%
The dataset is fully described in \cite{DiBdataset} and available at \cite{svanstrom_fredrik_2020_5500576}.
The videos and audios are recorded at locations in and around Halmstad Airport (IATA/ICAO code: HAD/ESMT), Gothenburg City Airport (GSE/ESGP) and Malmö Airport (MMX/ESMS).
Three different drones are used to collect and compose the dataset. These are of the following types: Hubsan H107D+, a small-sized first-person-view (FPV) drone; the high-performance DJI Phantom 4 Pro; and finally, the medium-sized kit drone DJI Flame Wheel. This can be built as a quadcopter (F450) or a hexacopter configuration (F550). The version used in this work is an F450 quadcopter. All three types can be seen in Figure~\ref{fig:db_drones}.
These drones differ a bit in size, with Hubsan H107D+ being the smallest, having a side length from motor to motor of 0.1 m. The Phantom 4 Pro and the DJI Flame Wheel F450 are larger with 0.3 and 0.4	m motor-to-motor side lengths, respectively.
To comply with regulations (drones must be flown within visual range), the dataset is recorded in daylight, even if the thermal infrared or acoustic sensors could be effective at night.

The videos and audio files are cut into ten-second segments to facilitate annotation. 
The acquisition was performed during the drastic reduction in air traffic due to the COVID19 pandemic. Therefore, to get a more comprehensive dataset, both in terms of aircraft types and sensor-to-target distances, the data is complemented with non-copyrighted material of \cite{[VIRTUALAIRFIELD]}. 
Overall, the dataset contains 90 audio clips and 650 videos (365 IR and 285 visible, of ten seconds each), with a total of 203328 annotated images (publicly available).
The thermal infrared videos have a resolution of 320$\times$256 pixels, and the visible videos have 640$\times$512.
The greatest sensor-to-target distance for a drone in the dataset is 200 m.
Audio files consist of 30 ten-second clips of each of the three output audio classes (Table~\ref{tab:output-classes}), and the amount of videos among the four output video classes is shown in Table~\ref{tab:db-stats}.
The background sound class contains general background sounds recorded outdoor in the typical deployment environment of the system, and also includes some clips of the sounds from the servos moving the pan/tilt platform.

\begin{table}[htb]
\caption{Distribution of the thermal infrared and visible videos. Each video has approximately 10 seconds. Reprinted with permission from \cite{drone20icpr}. \label{tab:db-stats}}

\begin{center}
\begin{tabular}{cccccccccc}


\multicolumn{1}{c}{} & \multicolumn{4}{c}{thermal infrared (365)}  & & \multicolumn{4}{c}{visible (285)} \\  \cline{2-5} \cline{7-10}

bin & airpl. & bird & drone & helicop.  & &  airpl. & bird & drone & helicop. \\ \hline
Close & 9 & 10 & 24  & 15  & &  17 & 10 & 21  & 27 \\
Medium & 25 & 23 & 94 & 20  & & 17 & 21 & 68 & 24 \\
Distant & 40 & 46 & 39 & 20  & & 25 & 20 & 25 & 10 \\ \hline

\end{tabular}

\end{center}

\end{table}
\normalsize

\begin{figure} [htb]
\centering
\includegraphics[width=0.7\textwidth]{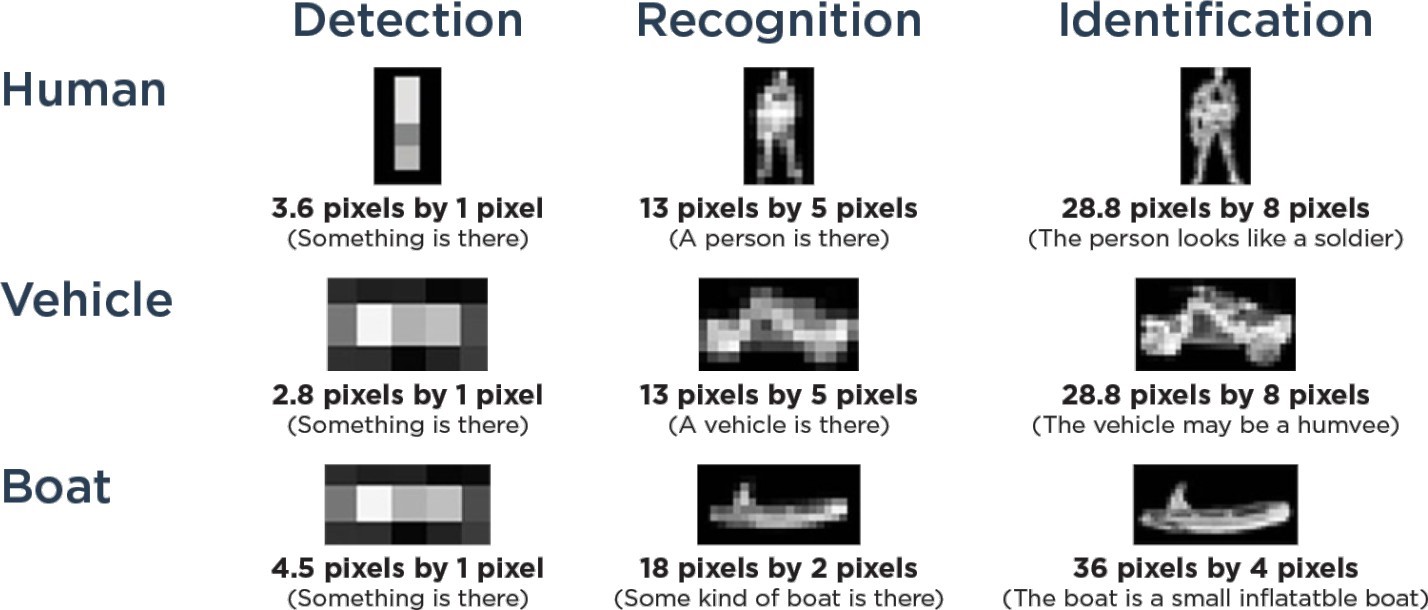}
\caption{The DRI requirements (from \cite{[DRI]}).}
\label{fig:DRI}
\end{figure}

The video part of the database is divided into three category bins: Close, Medium and Distant.
This is because one of our aims is to explore performance as a function of the sensor-to-target distance.
The borders between these bins are chosen to follow the industry-standard Detect, Recognize and Identify (DRI) requirements \cite{[DRI]}, building on the Johnson criteria \cite{[Chevalier16]}, as shown in Figure~\ref{fig:DRI}.
The Close bin is from 0 m to a distance where the target is 15 pixels wide (the requirement for `identification', e.g. explicitly telling the specific drone model, aircraft, helicopter, bird, etc.). The Medium bin is from where the target is from 15 down to 5 pixels (`recognition' of the target, e.g. a drone, an aircraft, a helicopter... albeit without the possibility of identifying the model), and the Distant bin is beyond that (`detection', e.g. there is something).

\begin{table}[htb]
\caption{Results with the thermal infrared sensor (confidence threshold and IoU requirement of 0.5). The average of the three F1-scores is 0.7601. Reprinted with permission from \cite{drone20icpr}. \label{tab:results-IRcam}}

\begin{center}
\begin{tabular}{cccccc}


\multicolumn{1}{c}{} & \multicolumn{5}{c}{distance bin: \textbf{CLOSE}}   \\ \cline{2-6}

 & airplane & bird & drone  & helicopter  &  average \\ \hline

\textbf{Precision} & 0.9197 & 0.7591 & 0.9159 & 0.9993& 0.8985 \\
\textbf{Recall} & 0.87367 & 0.85087 & 0.87907 & 0.87927 & 0.8706   \\ \hline
\textbf{F1-score} & & & & & 0.88447   \\
\hline

\multicolumn{6}{c}{} \\

\multicolumn{1}{c}{} & \multicolumn{5}{c}{distance bin: \textbf{MEDIUM}}   \\ \cline{2-6}

 & airplane & bird & drone  & helicopter  &  average \\ \hline

\textbf{Precision} &  0.82817 & 0.50637 & 0.89517 & 0.95547 & 0.7962  \\
\textbf{Recall} &  0.70397 & 0.70337 & 0.80347 & 0.83557 & 0.7615 \\ \hline
\textbf{F1-score} & & & & & 0.77857   \\ \hline

\multicolumn{6}{c}{} \\

\multicolumn{1}{c}{} & \multicolumn{5}{c}{distance bin: \textbf{DISTANT}}   \\ \cline{2-6}

 & airplane & bird & drone  & helicopter  &  average \\ \hline

\textbf{Precision} &  0.78227 & 0.61617 & 0.82787 & 0.79827 & 0.7561  \\
\textbf{Recall} &  0.40437 & 0.74317 & 0.48367 & 0.45647 & 0.5218 \\ \hline
\textbf{F1-score} & & & & & 0.61757   \\ \hline

\end{tabular}

\end{center}

\end{table}
\normalsize

\begin{table}[htb]
\caption{Results with the thermal infrared sensor (confidence threshold 0.8, IoU requirement 0.5). Table originally appearing in \cite{drone20thesis}. \label{tab:results-IRcam-th0.8}}

\begin{center}
\begin{tabular}{ccccc}

\multicolumn{1}{c}{} & \multicolumn{3}{c}{distance bin} & \multicolumn{1}{c}{Average}  \\ \cline{2-4}

 & close & medium & distant &  \\ \hline

\textbf{Precision} & 0.9985 & 0.9981 & 1.0000 & 0.9987\\
\textbf{Recall} & 0.2233 & 0.1120 & 0.0019 & 0.1124 \\ \hline

\end{tabular}

\end{center}

\end{table}
\normalsize

\section{Results}
\label{sect:results}

We provide evaluation results of the individual sensors in terms of precision, recall and F1-score.
Precision is the fraction of the total number of true positive detections. In other words, it measures how many of the detected objects are relevant.
Recall is the fraction of the total number of labelled samples in the positive class that are true positive. In other words, how many of the relevant objects are detected.
With the video detectors, these metrics can be obtained using the Matlab function \texttt{bboxPrecisionRecall}, which measures the accuracy of bounding box overlap between detected and ground truth boxes.
Since we also have confidence scores, the \texttt{evaluateDetectionPrecision} function can be used to plot precision curves as a function of the recall value.
After that, we perform sensor fusion experiments, a direction barely followed in the related literature \cite{Diamantidou19,Shi18,Unlu19}.
To carry out experiments, two disjoint sets of videos were created for training and testing (i.e. a video selected for the training set is not used in the testing set). 
The training set comprises 120 infrared and 120 visible clips (10 for each class and target bin per spectrum), resulting in 37428 infrared and 37519 visible frames.
The evaluation set comprises 60 infrared and 60 visible clips (5 for each class and target bin per spectrum), resulting in 18691 infrared and 18773 visible frames.
Since the duration of each clip is roughly the same ($\sim$10 seconds), the amount of samples per class and target bin is approximately balanced.
%
%
%
%
For the audio classifier, five 10-second clips from each output category are selected for evaluation, and the remaining clips for training. 
Since the audio classifier processes a one-second input buffer, the clips cut into that length, with an overlap of 0.5 seconds, resulting in 20 samples per clip.
This results in 297 clips in the evaluation set, 99 from each class.

In evaluating the YOLOv2 detector, it provides an array of class labels, detection confidence, and bounding boxes of detected objects.
Here, not only the classification label but also the placement of the bounding box must be taken under consideration.
When it produces multiple bounding boxes for the same object, only the strongest one is used, chosen as the box with the highest IoU (Intersection Over Union) with the annotations in the training data.
To assign a box to a class, it must also have a minimum IoU with the training object it is supposed to detect.
In related works, an IoU of 0.5 is usually employed \cite{Park17,Saqib17,Aker17}, although a lower IoU of 0.2 is used in \cite{Schumann17}.
In this work, we will stick to 0.5.
A threshold to the detection confidence can also be imposed, so bounding boxes with small confidence can be rejected, even if their IoU is above 0.5.

\begin{figure} [htb]
\centering
\includegraphics[width=0.6\textwidth]{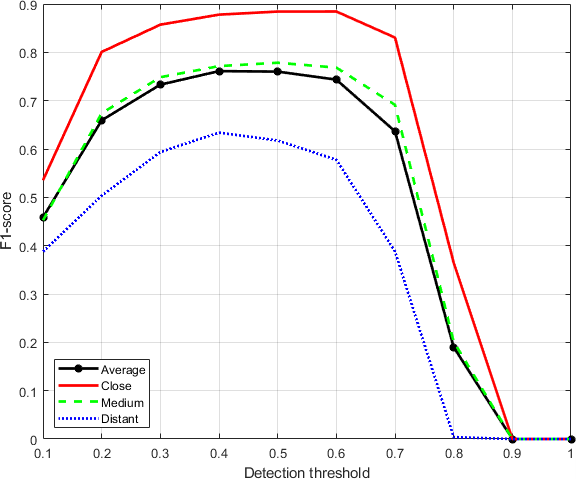}
\caption{F1-score with the thermal infrared sensor as a function of detection threshold, using all the 18691 images in the evaluation dataset. Picture originally appearing in \cite{drone20thesis}.}
\label{fig:results-IRcam-F1}
\end{figure}

\begin{figure} [htb]
\centering
\includegraphics[width=0.85\textwidth]{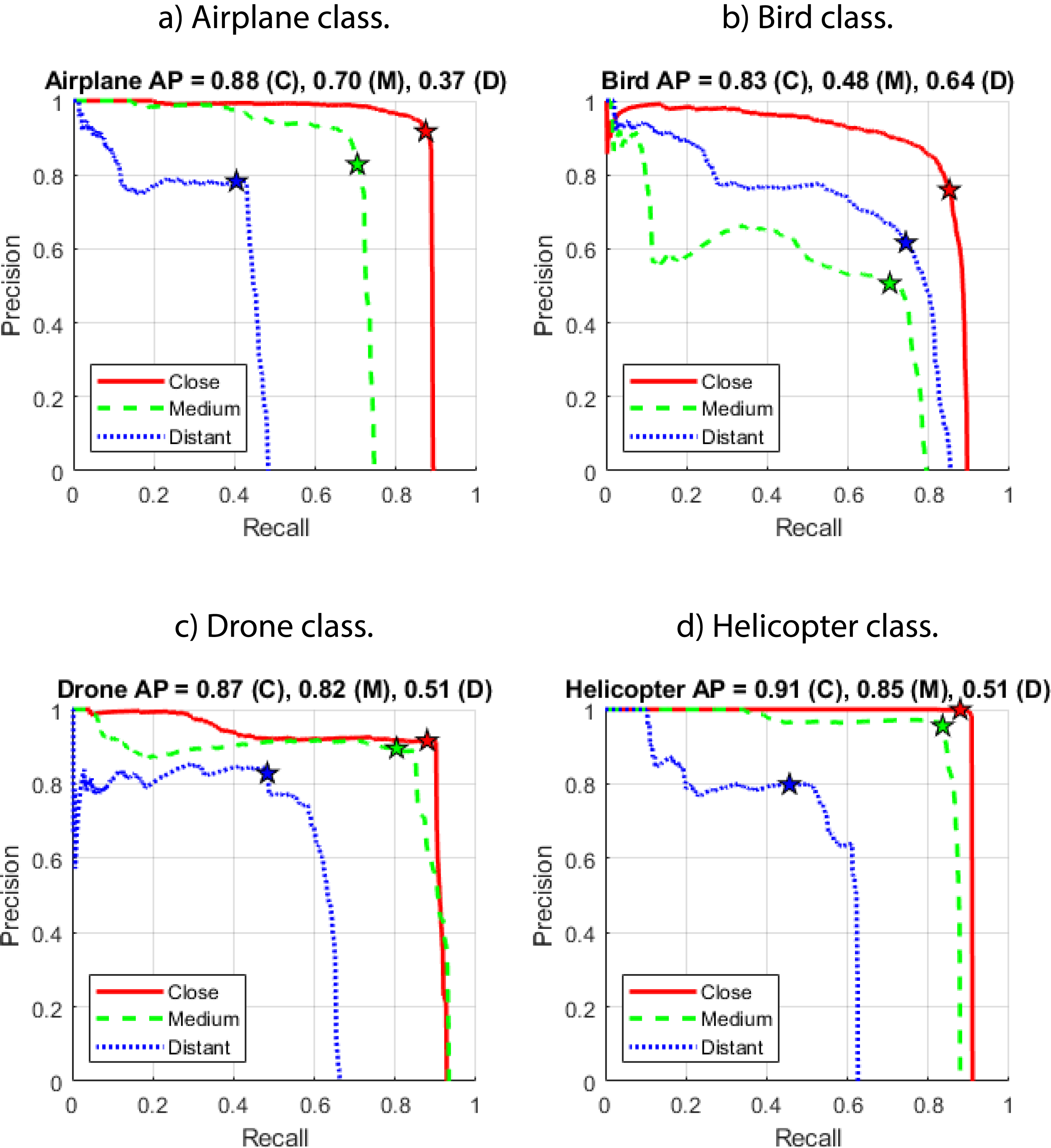}
\caption{Precision and recall curves with the thermal infrared sensor. The achieved values with a detection threshold of 0.5 are marked by stars. Pictures originally appearing in \cite{drone20thesis}.}
\label{fig:results-IRcam-PR}
\end{figure}

\subsection{Thermal Infrared Camera (IRcam)}

The precision, recall and F1-score of the IRcam worker detector with a confidence threshold set to 0.5 and an IoU requirement of 0.5 are shown in Table~\ref{tab:results-IRcam}.
We can observe that the precision and recall values are well balanced.
Altering the setting to a higher decision threshold of 0.8 leads to higher precision at the cost of a lower recall value, as shown in Table~\ref{tab:results-IRcam-th0.8}. The drop in recall with increasing sensor-to-target distance is also prominent.

To further explore the detection threshold setting, we run the evaluation with values from 0.1 up to 1.0 in steps of 0.1. The results in the form of an F1-score are shown in Figure~\ref{fig:results-IRcam-F1}.
Using not only the bounding boxes and class labels but also the confidence scores, the detector can be evaluated in the form of precision vs recall curves as well (Figure~\ref{fig:results-IRcam-PR}).
Note that the average precision results output from the Matlab \texttt{evaluateDetectionPrecision} function is defined as the area under the precision vs recall curve. Hence it is not the same as the actual average precision values of Table~\ref{tab:results-IRcam}.
To distinguish that we mean the area under the curve, we denote this as AP in Figure~\ref{fig:results-IRcam-PR}, just as in the original YOLO paper \cite{Redmon17}.
This is also  the definition used in \cite{Saqib17}, and to further adopt the notation of these papers, we denote the mean AP taken over all classes as the mAP, which is given in Table~\ref{tab:results-IRcam-mAP}.

\begin{table}[htb]
\caption{Mean values over all classes of the area under the precision vs recall curve (mAP) of the thermal infrared sensor. Table originally appearing in \cite{drone20thesis}. \label{tab:results-IRcam-mAP}}

\begin{center}
\begin{tabular}{ccccc}

\multicolumn{1}{c}{} & \multicolumn{3}{c}{distance bin} & \multicolumn{1}{c}{Average}  \\ \cline{2-4}

 & close & medium & distant &  \\ \hline

\textbf{mAP} & 0.8704 & 0.7150 & 0.5086 & 0.7097 \\ \hline

\end{tabular}

\end{center}

\end{table}
\normalsize

\begin{figure} [htb]
\centering
\includegraphics[width=0.85\textwidth]{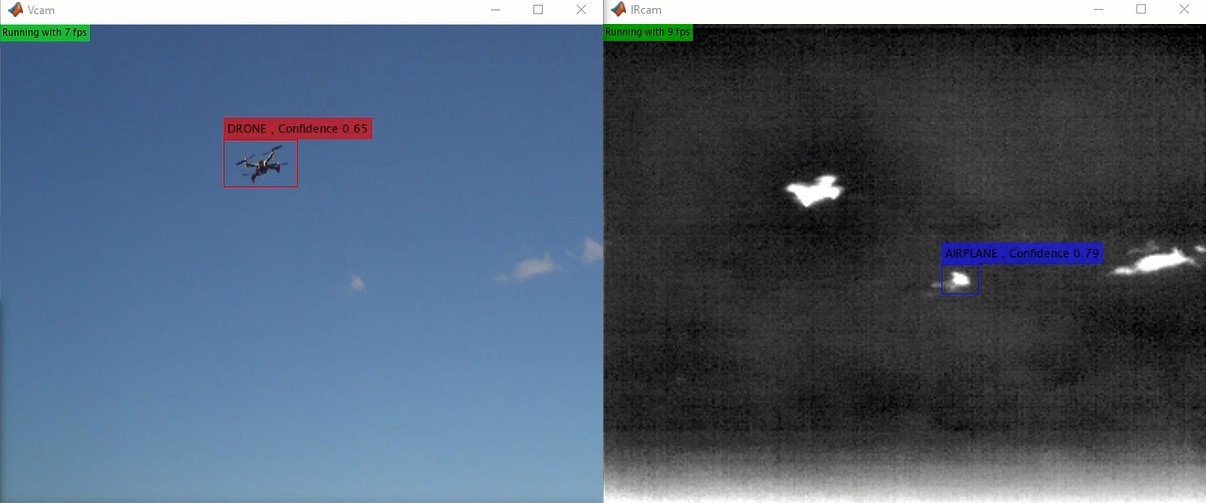}
\caption{A false target of the thermal infrared sensor caused by a small cloud lit by the sun. Pictures originally appearing in \cite{drone20thesis}.}
\label{fig:results-IRcam-false-target}
\end{figure}

The choice of the detection threshold will affect the achieved precision and recall values. The stars in Figure~\ref{fig:results-IRcam-PR} indicate the results with a detection threshold of 0.5 (as reported in Table~\ref{tab:results-IRcam}). We can conclude that such a threshold results in a balanced precision-recall combination near the top right edge of the respective curves. 
Compare this balanced behaviour to the  precision and recall values obtained with a detection threshold of 0.8 (Table~\ref{tab:results-IRcam-th0.8}).
From observations of the behaviour when running the drone detection system, we can also point out that a common source of false alarms of the thermal infrared sensor is small clouds and edges of large clouds lit up by the sun. An example of this can be seen in Figure~\ref{fig:results-IRcam-false-target}.

In comparing our work with other research using thermal infrared sensors, results can be found in \cite{ANDRASI2017183}. 
However, the sensor used in that article to detect drones (up to a distance of 100 m) has a resolution of just 80$\times$60 pixels, and the system does not involve any form of machine learning feature (the task is done by a human looking at the output video stream). 
%
%
In \cite{wang_chen_choi_kuo_2019}, curves for the precision and recall of a machine learning-based thermal detector are presented. It is stated that the video clips used for training and evaluation have a frame resolution of 1920$\times$1080. Unfortunately, the paper fails to mention if this is also the resolution of the sensor. Neither is the input size of the detection network specified in detail, other than the images are re-scaled so that the shorter side has 600 pixels.
The most substantial results to relate to in \cite{wang_chen_choi_kuo_2019} is that since the system also contains a video camera with the same image size as the thermal one, the authors can conclude that the thermal drone detector performs over 8\% better than the video detector.

\subsection{Video Camera (Vcam)}

The results of the previous sub-section will be compared to the results of the video detector in what follows.
To enable comparison, the same methods and settings as the infrared camera are used here, leading to the precision, recall and F1-score results of Table~\ref{tab:results-Vcam} with a confidence threshold and IoU requirement of 0.5.
These results differ no more than 3\% from the results of the thermal infrared detector. Recall that the input layers of the YOLOv2 detectors are different, and hence, the resolution of the visible sensor is 1.625 higher than the thermal infrared detector (416$\times$416 vs 256$\times$256).
So even with a lower resolution and using images in grey-scale and not in colour, the thermal infrared sensor performs as well as the visible one. 
This is in line with \cite{ANDRASI2017183}, where the infrared detector outperforms the visible one when the image size is the same, although in our case, this happens even with a smaller image size.

\begin{table}[htb]

\caption{Results with the visible camera (confidence threshold and IoU requirement of 0.5). The average of the three F1-scores is 0.7849. Reprinted with permission from \cite{drone20icpr}. \label{tab:results-Vcam}}
\begin{center}
\begin{tabular}{cccccc}


\multicolumn{1}{c}{} & \multicolumn{5}{c}{distance bin: CLOSE}   \\ \cline{2-6}

 & airplane & bird & drone  & helicopter  &  average \\ \hline

\textbf{Precision} & 0.8989 & 0.8284 & 0.8283 & 0.9225 & 0.8695 \\
\textbf{Recall} & 0.7355 & 0.7949 & 0.9536 & 0.9832 & 0.8668 \\ \hline
\textbf{F1-score} & & & & & 0.8682   \\
\hline

\multicolumn{6}{c}{} \\

\multicolumn{1}{c}{} & \multicolumn{5}{c}{distance bin: MEDIUM}   \\ \cline{2-6}

 & airplane & bird & drone  & helicopter  &  average \\ \hline

\textbf{Precision} & 0.8391 & 0.7186 & 0.7710 & 0.9680 & 0.8242 \\
\textbf{Recall} & 0.7306 & 0.7830 & 0.7987 & 0.7526 & 0.7662 \\ \hline
\textbf{F1-score} & & & & & 0.7942   \\ \hline

\multicolumn{6}{c}{} \\

\multicolumn{1}{c}{} & \multicolumn{5}{c}{distance bin: DISTANT}   \\ \cline{2-6}

 & airplane & bird & drone  & helicopter  &  average \\ \hline

\textbf{Precision} & 0.7726 & 0.6479 & 0.8378 & 0.6631 & 0.7303 \\
\textbf{Recall} & 0.7785 & 0.7841 & 0.5519 & 0.5171 & 0.6579 \\ \hline
\textbf{F1-score} & & & & & 0.6922   \\ \hline

\end{tabular}

\end{center}

\end{table}
\normalsize

\begin{figure} [htb]
\centering
\includegraphics[width=0.6\textwidth]{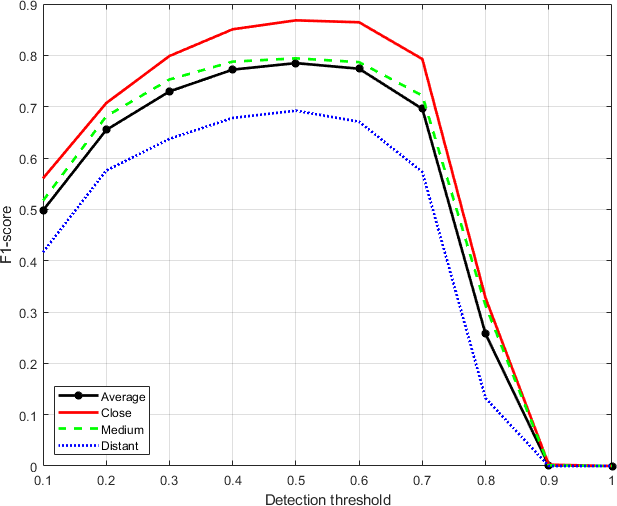}
\caption{F1-score with the visible camera as a function of detection threshold, using all the 18773 images in the evaluation dataset. Picture originally appearing in \cite{drone20thesis}.}
\label{fig:results-Vcam-F1}
\end{figure}

\begin{figure} [htb]
\centering
\includegraphics[width=0.85\textwidth]{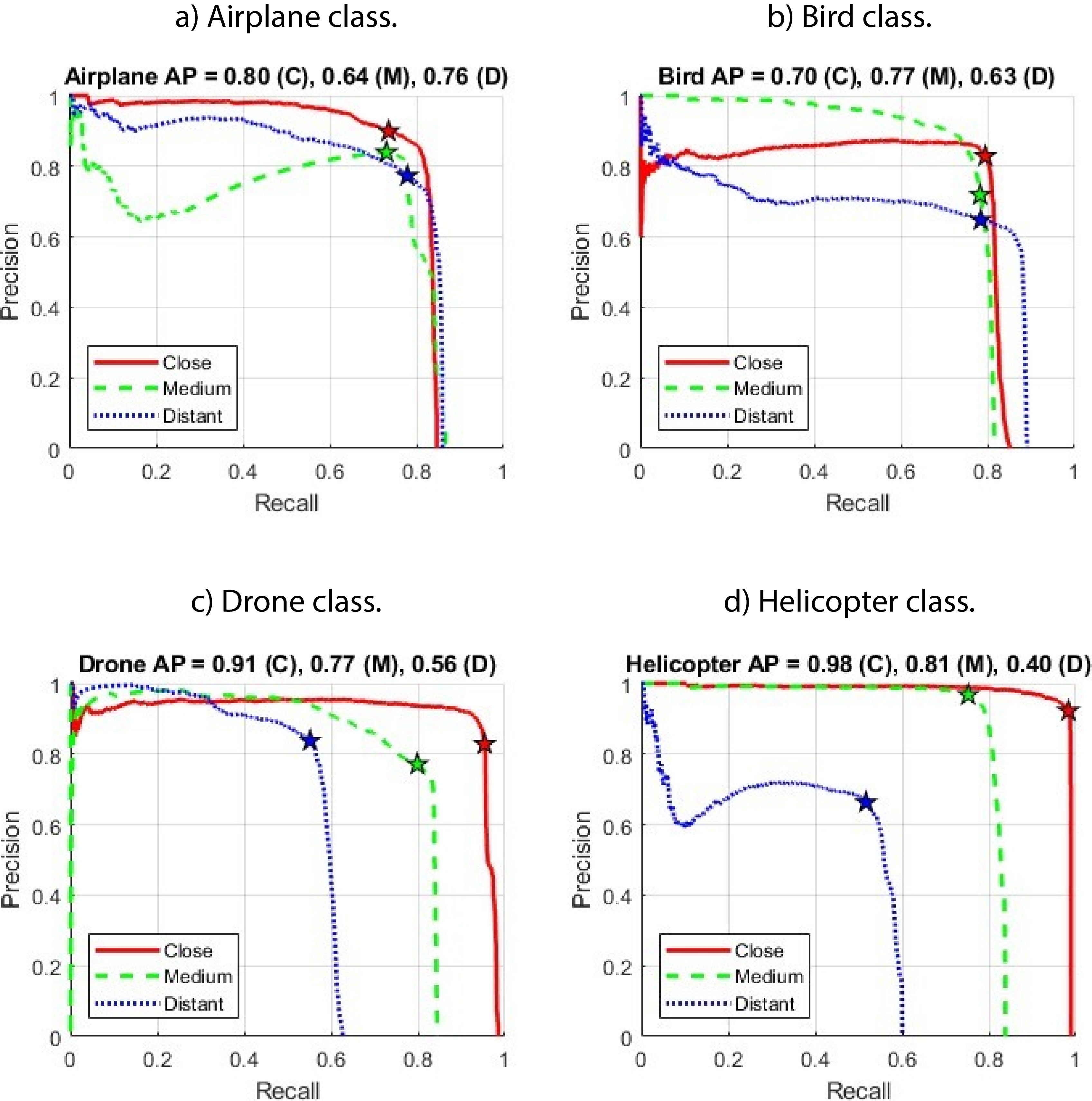}
\caption{Precision and recall curves with the visible camera. The achieved values with a detection threshold of 0.5 are marked by stars. Pictures originally appearing in \cite{drone20thesis}.}
\label{fig:results-Vcam-PR}
\end{figure}

Just as for the thermal infrared camera, we can also explore the effects of the detection threshold setting. This can be seen in Figure~\ref{fig:results-Vcam-F1}.
The precision vs recall curves of the video camera images for the different target classes and distance bins are shown in Figure~\ref{fig:results-Vcam-PR}.
In a similar vein, a detection threshold of 0.5 results in a balanced precision-recall combination near the top right edge of the respective curves.
Notably, when inspecting the precision-recall curves, the video camera detector performs outstandingly when it comes to distant airplanes. This has its explanation in that such targets often presents a very large signature consisting not only of the airplane itself but also contrails behind it.
Also, calculating the mAP from these results, we obtain Table~\ref{tab:results-Vcam-mAP}, with an average of 0.7261.
Once again, it is not far from the 0.7097 mAP of the thermal infrared detector of the previous sub-section.

\begin{table}[htb]
\caption{Mean values over all classes of the area under the precision vs recall curve (mAP) of the visible camera. Table originally appearing in \cite{drone20thesis}. \label{tab:results-Vcam-mAP}}

\begin{center}
\begin{tabular}{ccccc}

\multicolumn{1}{c}{} & \multicolumn{3}{c}{distance bin} & \multicolumn{1}{c}{Average}  \\ \cline{2-4}

 & close & medium & distant &  \\ \hline

\textbf{mAP} & 0.8474 & 0.7477 & 0.5883 & 0.7261 \\ \hline

\end{tabular}

\end{center}

\end{table}
\normalsize

\begin{figure} [htb]
\centering
\includegraphics[width=0.85\textwidth]{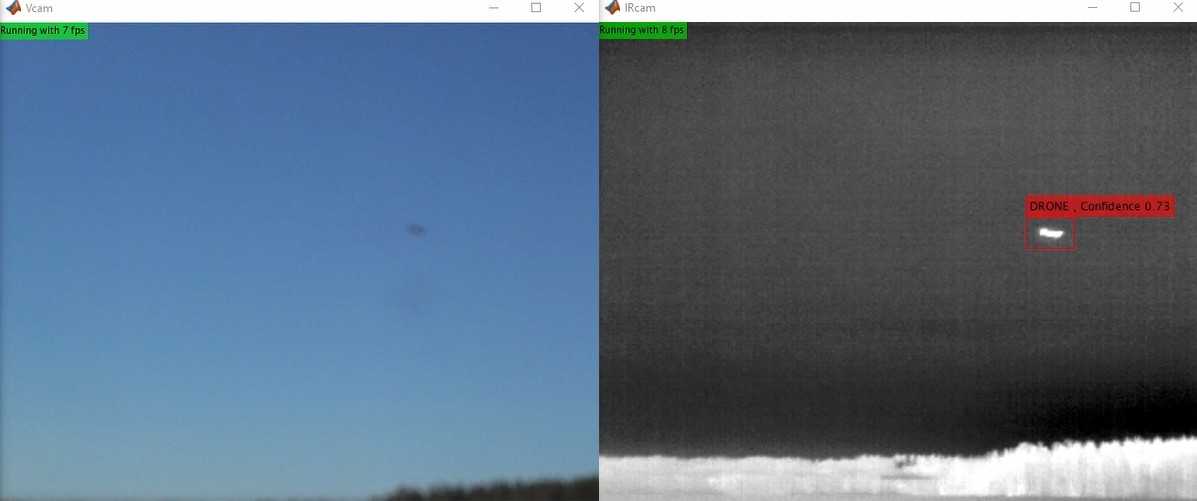}
\caption{A drone is detected only by the thermal infrared sensor since the auto-focus of the video camera is wrongly set. Pictures originally appearing in \cite{drone20thesis}.}
\label{fig:results-Vcam-autofocus-fail}
\end{figure}

The most frequent problem with the video detector is the auto-focus feature of the video camera. For this type of sensor, clear skies are not the ideal weather, but rather scenery with objects that can help the camera to set the focus correctly. However, note that this fact is not heavily affecting the evaluation results of the visible camera detector performance, as presented above, since only videos where the objects are seen clearly, and hence are possible to annotate, are used.
Figure~\ref{fig:results-Vcam-autofocus-fail} shows an example of when the focus is set wrongly so that only the thermal infrared worker detects the drone.
This issue justifies the multi-sensor approach followed here. Also, cameras not affected by this issue (such as the thermal infrared or the fish-eye) could be used to aid the focus of the video camera. 

%
Comparing the results to other papers, our results are ahead of what is presented in \cite{Saqib17}, where a mAP of 0.66 was achieved, albeit using a detector with drones as the only output class and giving no information about the sensor-to-target distances.
Also, we see that the YOLOv2 detector in \cite{Park17} achieves an F1-score of 0.728 with the same detection threshold and IoU-requirement. This F1-score is just below the results of the thermal infrared and video camera workers.
However, one notable difference lies in that the detector in \cite{Park17} has only one output class. This fact could confirm the doctrine of this work, i.e. that the detectors should also be trained to recognize objects easily confused for being drones. Unfortunately, there is no notation of the sensor-to-target distance other than that ``75\% of the drones have widths smaller than 100 pixels''. Since the authors implement an original YOLOv2 model from darknet, it is assumed that the input size of the detector is 416$\times$416 pixels.
A YOLOv2 architecture with an input size of 480$\times$480 pixels is implemented in \cite{Aker17}. The detector has two output classes, birds and drones. Based on the presented precision-recall curve, a precision and recall of 0.9 can be achieved simultaneously.
To summarize this comparison, we provide the performance of the thermal infrared and video camera detectors together with the reported comparable results in Table~\ref{tab:results-IRcam-Vcam-SOA}. The table also shows the output classes used.

\begin{table}[htb]
\caption{Results from the related works and the thermal infrared and video camera detectors. A = Airplane, B = Bird, D = Drone and H = Helicopter. Table originally appearing in \cite{drone20thesis}. \label{tab:results-IRcam-Vcam-SOA}}

\begin{center}
\begin{tabular}{ccccccc}

\multicolumn{1}{c}{Work} & \multicolumn{1}{c}{Sensor} & \multicolumn{4}{c}{Results} & \multicolumn{1}{c}{Classes}   \\ \cline{3-6}

 & & Precision & Recall & F1-score & mAP &  \\ \hline


 
Park et al. \cite{Park17} & Visible &  &  & 0.73 & & D  \\ 

Liu  et al. \cite{Liu_2018} & Visible & 0.99 & 0.80  &  & & A, D, H   \\ 

Saqib et al. \cite{Saqib17} & Visible &  &  &  & 0.66 & D   \\ 

Aker and Kalkan. \cite{Aker17} & Visible & $\sim$0.9 & $\sim$0.9 &  &  & B, D  \\ 

\textbf{thermal infrared} & Thermal & 0.82 & 0.72 & 0.76 & 0.71 & A, B, D, H  \\ 

\textbf{video camera} & Visible & 0.81 & 0.76 & 0.78 & 0.73 & A, B, D, H   \\  

\hline

\end{tabular}

\end{center}

\end{table}
\normalsize

\subsection{Fish-Eye Camera Motion Detector (Fcam)}

The fish-eye lens camera is included to be able to cover a larger airspace volume than covered by just the field of view of the thermal infrared and video camera. However, the drone detection system does not totally rely on the fish-eye worker to detect objects of interest since the system also includes the search programs moving the pan/tilt platform when no other targets are detected. The search programs can easily be turned on or off using the control panel of the GUI.

\begin{figure} [htb]
\centering
\includegraphics[width=0.85\textwidth]{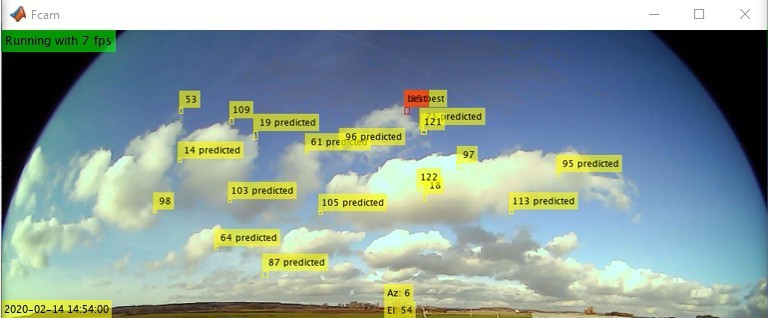}
\caption{False targets in a fish-eye camera image. Picture originally appearing in \cite{drone20thesis}.}
\label{fig:results-Fcam-false-targets}
\end{figure}

It was initially observed that the foreground/background-based motion detector of the fish-eye worker was sensitive to objects moving with the wind, as shown in Figure~\ref{fig:results-Fcam-false-targets}.
The false targets have been mitigated by extensive tuning of the image processing operations, the GMM foreground detector \cite{gmmmatlab}, the blob analysis settings and the parameters of the multi-object Kalman filter tracker from \cite{kalmanmatlab} as follows. All other parameters or computation methods not mentioned here are left at the default values described in \cite{gmmmatlab},\cite{kalmanmatlab}.

\begin{itemize}

\item Some of the most critical points in this were first to remove the \texttt{imclose}\footnote{Morphological closing.} and \texttt{imfill}\footnote{Flood-fill of holes, i.e. background pixels that cannot be reached by filling in the background from the edge of the image.} functions that were initially implemented after the \texttt{imopen}\footnote{Morphological opening.} function in the image processing operations.

\item To only detect small targets, in the blob analysis settings of the \texttt{BlobAnalysis} function, a maximum blob area of 1000 pixels was implemented.



\item The parameters of the \texttt{ForegroundDetector} function were changed so that the model would adapt faster to changing conditions, and hence react quicker. 
The number of initial video frames for training the background model was reduced from 150 (default) to 10, and the learning rate for parameter updates increased from 0.005 (default) to 0.05. 
Furthermore, to reduce false alarms, the threshold to determine the background model (i.e. minimum possibility for pixels to be considered background values) was increased from 0.7 (default) to 0.85.
The number of Gaussian modes in the mixture model is left to the default value of 5.

%

\item Tuning the parameters of the Kalman filter multi-object tracker was also important. These were altered to make the tracker slower to start new tracks and quicker to terminate them if no moving objects were present at the predicted positions.
The motion model in the \texttt{configureKalmanFilter} function was chosen to be of the type ``constant velocity''.
The initial location of unassigned objects is given by the centroid computed by the foreground detector.
The initial estimate error was changed from [200, 50] (default) to [200, 200]. This specifies the variance of the initial estimates of location and velocity of the tracked object, with the initial state estimation error covariance matrix built as 2$\times$2 diagonal matrix with these two values on the main diagonal.
Larger values here help the filter to adapt to the detection results faster. %
This affect the first few detections, after which the estimate error is obtained from the noise and input data. The function assumes a zero initial velocity at the position of the initial location.
The motion noise was changed from [100, 25] (default) to [50, 50], with the process noise covariance matrix built as a 2$\times$2 diagonal matrix with these two values on the main diagonal. These values specify the tolerance (variance) for the deviation on location and velocity, compensating for the difference between the actual motion and that of the constant velocity model.
Here the changes were made since we must assume that the initial estimate error and the motion noise are uniform in x- and y-dimensions. 
The measurement noise covariance was increased from 10 (default) to 100, specifying the variance inaccuracy of the detected location.
Increasing this value enables the Kalman filter to remove more noise from the detections. This parameter was changed by trial and error.
Both the values of the motion noise and measurement noise stay constant.



\end{itemize}

%
%

%

\begin{figure} [htb]
\centering
\includegraphics[width=0.95\textwidth]{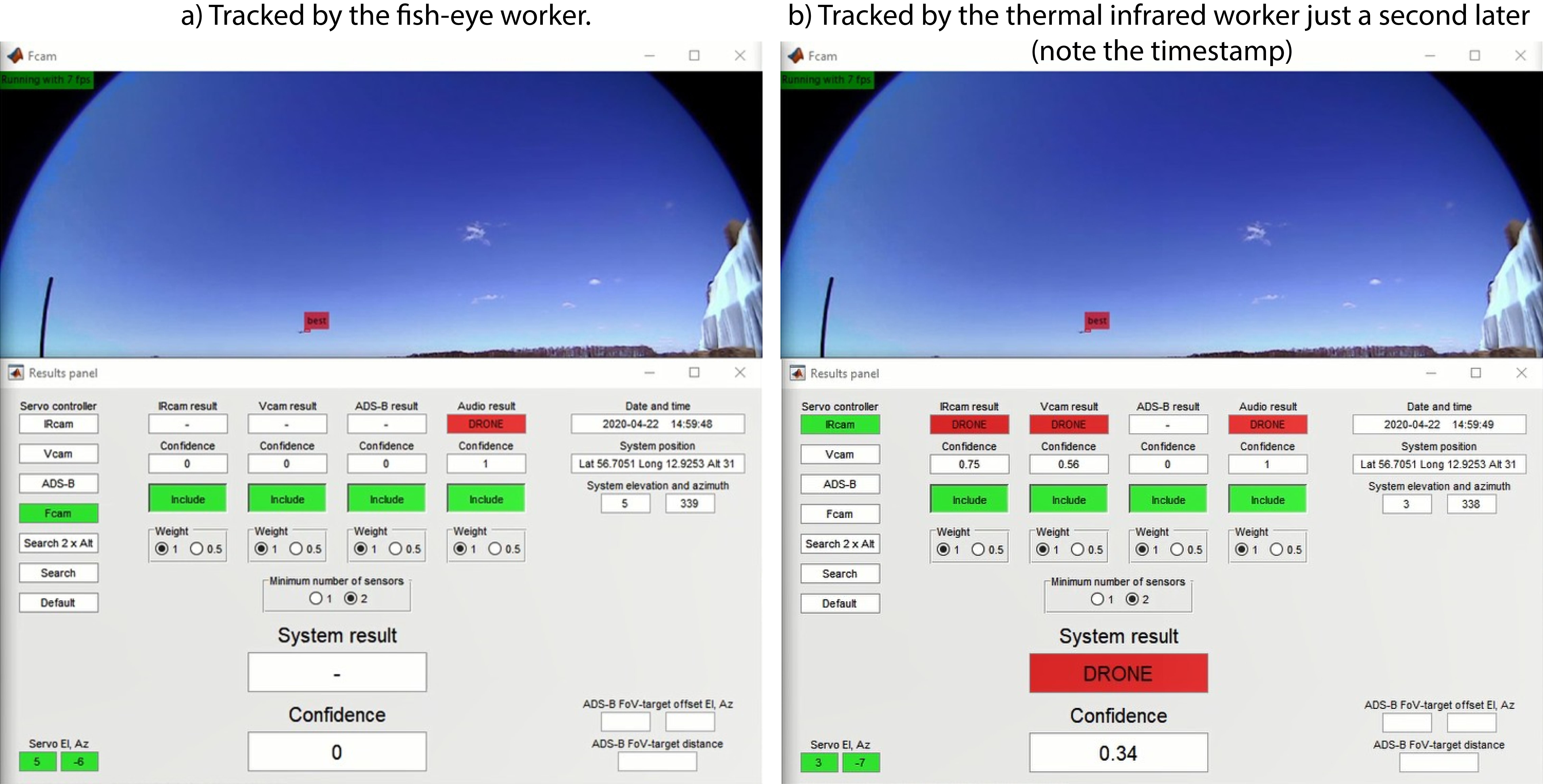}
\caption{Tracking of a drone with different workers. Pictures originally appearing in \cite{drone20thesis}.
}
\label{fig:fig36-37}
\end{figure}

With an image size of 1024$\times$384, the fish-eye worker moving object detector and tracker has been found, during the evaluation sessions, to be an effective way to assign objects to the pan/tilt platform up to a distance of 50 m against drones. 
Beyond this distance, the drone appears 
so small (in pixels) that it is deleted by the \texttt{imopen} function.
The maximum resolution of the fish-eye camera employed is 3264$\times$2448, so a greater detection range should theoretically be achievable. However, using a higher resolution is also more demanding in computational resources, leading to a reduction in the FPS performance for the fish-eye worker and the other workers. Since the fish-eye lens camera is also complemented by the search program, where the pan/tilt platform can be controlled by the output of the other workers, the choice has been made to settle with this resolution and the limited detection range that follows from this.
Figure~\ref{fig:fig36-37}a shows a drone tracked by the fish-eye worker. At this moment, the pan/tilt platform is controlled by the output of the fish-eye worker. Just a moment later, as seen in Figure~\ref{fig:fig36-37}b, the drone is detected by the thermal infrared and video camera workers, and the thermal infrared worker output therefore controls the pan/tilt platform.

\subsection{Microphone}

The precision and recall results of the different classes with this detector are shown in Table~\ref{tab:results-audio}. 
Against a drone, the practical range of the acoustic sensor is 35-45 m, depending on how the drone is flying. This is in parity with the 50 m of \cite{Park15}, but far from the 160 m against a F450 drone reported in
\cite{Busset15} with its much more complex microphone configuration (a 120 elements microphone array).

\begin{table}[htb]
\caption{Results with the audio detector. Reprinted with permission from \cite{drone20icpr}. \label{tab:results-audio}}

\begin{center}
\begin{tabular}{ccccc}


\multicolumn{5}{c}{}    \\ \cline{2-5}

 & drone & helicopter & background  & average \\ \hline

\textbf{Precision} & 0.9694 & 0.8482 & 0.9885 & 0.9354 \\
\textbf{Recall} & 0.9596 & 0.9596 & 0.8687 & 0.9293 \\ \hline
\textbf{F1-score} & & & & 0.9323   \\
\hline

\end{tabular}

\end{center}

\end{table}
\normalsize

Our average F1-score is 0.9323, which is higher compared to \cite{Jeon17}, which also uses MFCC features. Out of the three classifier types tested in \cite{Jeon17}, the one comprising a LSTM-RNN performs the best, with a F1-score of 0.6984. The classification problem in that paper is binary (drone or background).
Another paper applying MFCC-features and Support Vector Machines (SVM) as classifier is \cite{Bernardini17}, with a precision of 0.983. Five output classes are used (drone, nature daytime, crowd, train passing and street with traffic), and the classification is based on a one-against-one strategy. Hence ten binary SVM classifiers are implemented. 
The final output label is then computed using the max-wins voting principle. 
The paper \cite{Kim17}, on the other hand, applies the Fast Fourier Transform (FFT) to extract features from the audio signals.
The results of the audio detector, together with results reported in other studies, are summarized in Table~\ref{tab:results-audio-SOA}.

\begin{table}[htb]
\caption{Results from the related works and the audio detector. D = Drone, H = Helicopter, BG = Background. The work \cite{Bernardini17} defines four background classes: nature daytime, crowd, train passing and street with traffic. Table originally appearing in \cite{drone20thesis}. \label{tab:results-audio-SOA}}

\begin{center}
\begin{tabular}{ccccc}

\multicolumn{1}{c}{Work} & \multicolumn{3}{c}{Results} & \multicolumn{1}{c}{Classes}   \\ \cline{2-4}

 & Precision & Recall & F1-score &   \\ \hline

Kim et al. \cite{Kim17} & 0.88  & 0.83 & 0.85  &  D, BG  \\ 

Jeon et al. \cite{Jeon17} & 0.55 & 0.96 & 0.70  &  D, BG  \\ 

Bernardi et al. \cite{Bernardini17} & 0.98  &  &   &  D, BG  \\ 

\textbf{audio worker} & 0.94 & 0.93 & 0.93 &  D, H, BG  \\ 

\hline

\end{tabular}

\end{center}

\end{table}
\normalsize

\subsection{Radar Module}

From the datasheet of the K-MD2 \cite{RFbeam}, we have that it can detect a person with a Radar Cross Section (RCS) of 1 m$^2$ up to a distance of 100 m. Since we have from \cite{Patel18} that the RCS of the F450 drone is 0.02 m$^2$, it is straightforward to calculate that, theoretically, the F450 should be possible to detect up to a distance of $100 \times \sqrt[4]{{0.02/1}} = 37.6$ m. 
Furthermore, given that the micro-doppler echoes from the rotors are 20 dB below that of the drone body, these should be detectable up to a distance of $100 \times \sqrt[4]{{0.02/1 \times 100}} = 11.9$ m.

\begin{figure} [htb]
\centering
\includegraphics[width=0.6\textwidth]{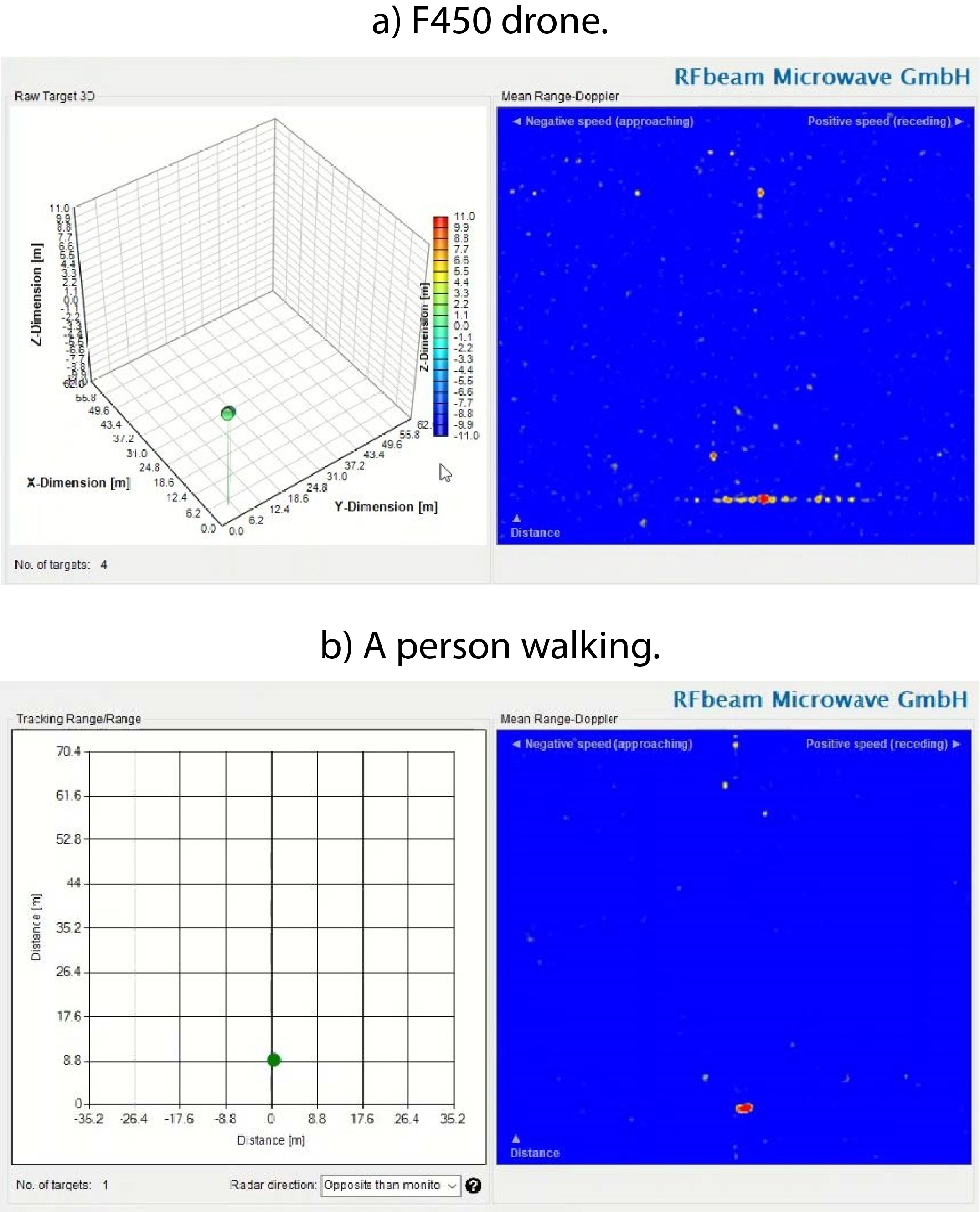}
\caption{Micro-doppler signature of different elements in front of the radar module. Pictures originally appearing in \cite{drone20thesis}.
}
\label{fig:fig40-41}
\end{figure}

We have observed that the F450 drone is detected and tracked by the K-MD2 up to a maximum distance of 24 m. This is, however, the maximum recorded distance, and it is observed that the drone is generally detected up to a distance of 18 m.
Due to the short practical detection range of the radar module, it is not included in the drone detection system of this work.
The micro-doppler signature can also be detected at short distances, as shown in Figure~\ref{fig:fig40-41}a. Corresponding range-doppler plots showing the micro-doppler of flying drones can be found in \cite{Drozdowicz16} and \cite{Rahman18} as well.
Compared to the echo of a person walking in front of the radar, as we can see in Figure~\ref{fig:fig40-41}b, no such micro-doppler signature is present in that case.

\subsection{Sensor Fusion}

We investigated the early or late sensor fusion choice based on whether the sensor data is fused before or after the detection element.
By early sensor fusion, we mean to fuse the images from the thermal infrared and video cameras before classification, either the raw images (sensor level fusion), or features extracted from them (feature level). 
%
%
Late sensor fusion will, in this case, be to combine the output decision from the separate detectors running on each camera stream, weighted by the confidence score of each sensor (decision level fusion). Other existing late fusion approaches entail combining the confidence scores (score level fusion) or the ranked class output of each system (rank level) \cite{FIERREZ201857}.

When investigating early sensor fusion at the raw image level in this research, the pixel-to-pixel matching of the images was the biggest issue. Even if this is possible in a static scenario, it turned out to be an unfeasible solution against moving objects with the available equipment due to the small but still noticeable difference in latency between the cameras. 
An interesting early sensor fusion method is also found in \cite{Unlu19} where the image from the narrow-angle camera is inserted into the image from the wide-angle camera and then processed by a single YOLO detector. It is unclear how they avoid situations when the inserted image obscures the object found in the wide-angle image.
In \cite{Diamantidou19}, they implemented early sensor fusion by concatenating feature vectors extracted from three different sensors (visible, thermal and 2D-radar), which are fed to a trained multilayer perceptron (MLP) classifier.
To do so here would likely require much more training data, not only on an individual sensor level but especially on a system level. Such amounts of data have not been possible to achieve within the scope of this work. 

\begin{figure} [htb]
\centering
\includegraphics[width=0.45\textwidth]{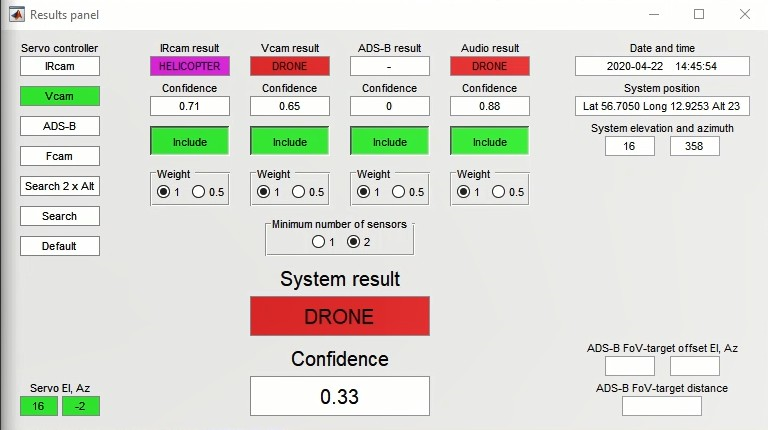}
\includegraphics[width=0.45\textwidth]{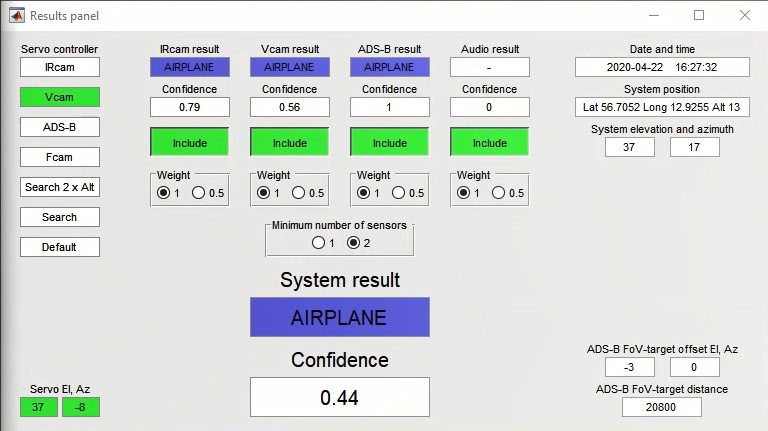}
\caption{Two examples of sensor fusion results. Pictures were originally appearing in \cite{drone20thesis} and published in \cite{drone20icpr}. Reprinted with permission.}
\label{fig:fusion-options}
\end{figure}

The mentioned issues are the motives for implementing non-trained late sensor fusion in this work.
The sensor fusion implemented consists of utilizing the class outputs and the confidence scores of the available sensors in a weighed manner, smoothing their result over time.
A weighted fusion approach is shown to be more robust compared to other non-trained techniques such as voting, majority rule, the arithmetic combination of confidences (via, e.g. mean, median or product), or taking the most confident classifier (max rule) \cite{FIERREZ201857}.
To carry out the fusion, every time the main script polls the worker’s queues, it puts the results in a 4$\times$4 matrix, organized so that each class is a column and each sensor is a row. 
The matrix values depend not only on the class label and the confidence but also on the setting of which sensors to include and the weight of the specific sensor, i.e. how much we trust it at the moment.
A new 1$\times$4 matrix is then formed by column-wise sum.
This array is in turn placed as a new row in a 10$\times$4 first-in-first-out time-smoothing matrix. Since we have close to ten FPS from the workers, this will be the results of approximately the last second. Once again, the 10 columns are summarized into a 1$\times$4 matrix, and the column with the highest value will be the output system class.
The system output confidence is calculated by normalizing the value found to be highest over the number of sensors included at the moment. An additional condition before presenting the system result is that the minimum number of sensors that detects any object must fulfil the GUI setting, as shown in Figure~\ref{fig:fusion-options}.
The figure also shows an example of how sensor fusion is enabled in the GUI, including the weight choice for each sensor and the minimum number required.
With such a dynamical setting, it is possible to use not only the OR-function, as in \cite{Shi18}, but more sophisticated variants by varying the number of sensors included and required for detection, including their weights.

\begin{figure} [htb]
\centering
\includegraphics[width=0.95\textwidth]{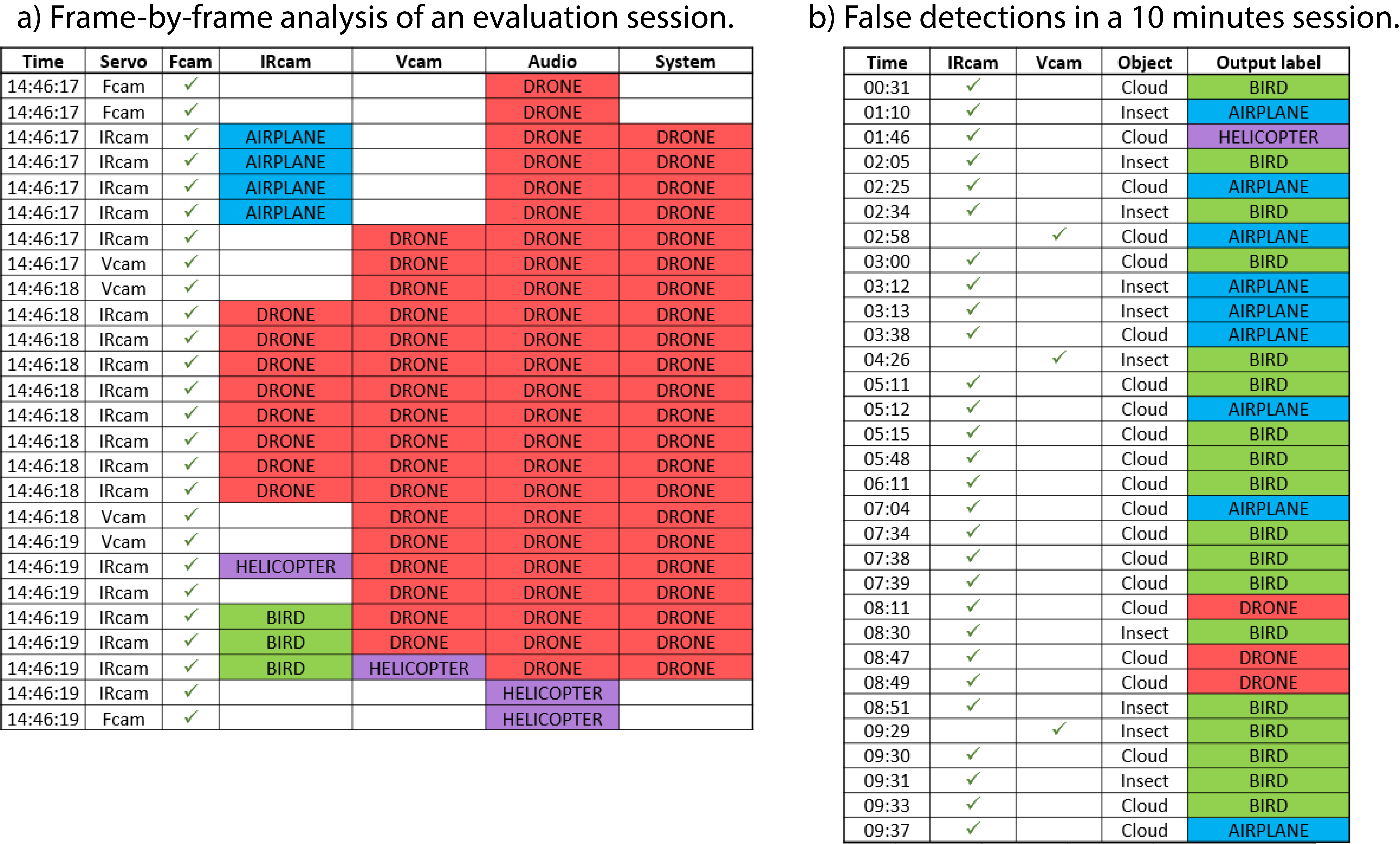}
\caption{Evaluating the efficiency the sensor fusion.
Pictures were originally appearing in \cite{drone20thesis} and published in \cite{drone20icpr}. Reprinted with permission.
}
\label{fig:fusion-results1}
\end{figure}

Evaluating the fusion in operational conditions was more challenging than expected since this research was mostly done during the COVID19 pandemic. 
Regular flights at the airports considered in this work were cancelled, hence the possibility for a thorough system evaluation against airplanes.
Using the screen recorder, it was possible to do a frame-by-frame analysis of a typical drone detection, as shown in Figure~\ref{fig:fusion-results1}a.
The ``servo'' column indicates the current servo controlling sensor and the ``Fcam'' column specifies if the fish-eye camera motion detector is tracking the drone. 
The respective class output labels of each worker are shown in the rest of the columns. Note that the system output is more stable and lasts for more frames than the thermal infrared and video camera individually, indicating the benefit of the sensor fusion. 
Since there is no information from the ADS-B receiver in this case, that column has been omitted from the table.
Figure~\ref{fig:system_GUI} above is the third frame from 14:46:18. As it can be seen, the thermal infrared, video camera, and audio workers detect and classify the drone correctly. The fish-eye camera worker also tracks the drone, and the thermal infrared worker controls the pan/tilt platform.

To measure the efficiency of the sensor fusion, we can consider occasions such as the one described in Figure~\ref{fig:fusion-results1}a as a \textit{detection opportunity}. If we define this to be when the drone is continuously observable in the field of view of the thermal infrared and video cameras, and hence possible for the system (including the audio classifier) to analyse and track, we have carried out 73 such opportunities in the screen recordings from the evaluation sessions.
The duration of the individual detection opportunities is from just fractions of a second up to 28 seconds. This is highly dependent on how the drone is flying and whether the system can track the drone. We can see that Figure~\ref{fig:fusion-results1}a describes the frame-by-frame analysis of a detection opportunity lasting for three seconds.
Comparing the system results after the sensor fusion with the output from the respective sensors, we can observe that the system outputs a drone classification at some time in 78\% of the detection opportunities. Closest to this is the performance of the video camera detector that outputs a drone classification in 67\% of the opportunities.

\begin{figure} [htb]
\centering
\includegraphics[width=0.7\textwidth]{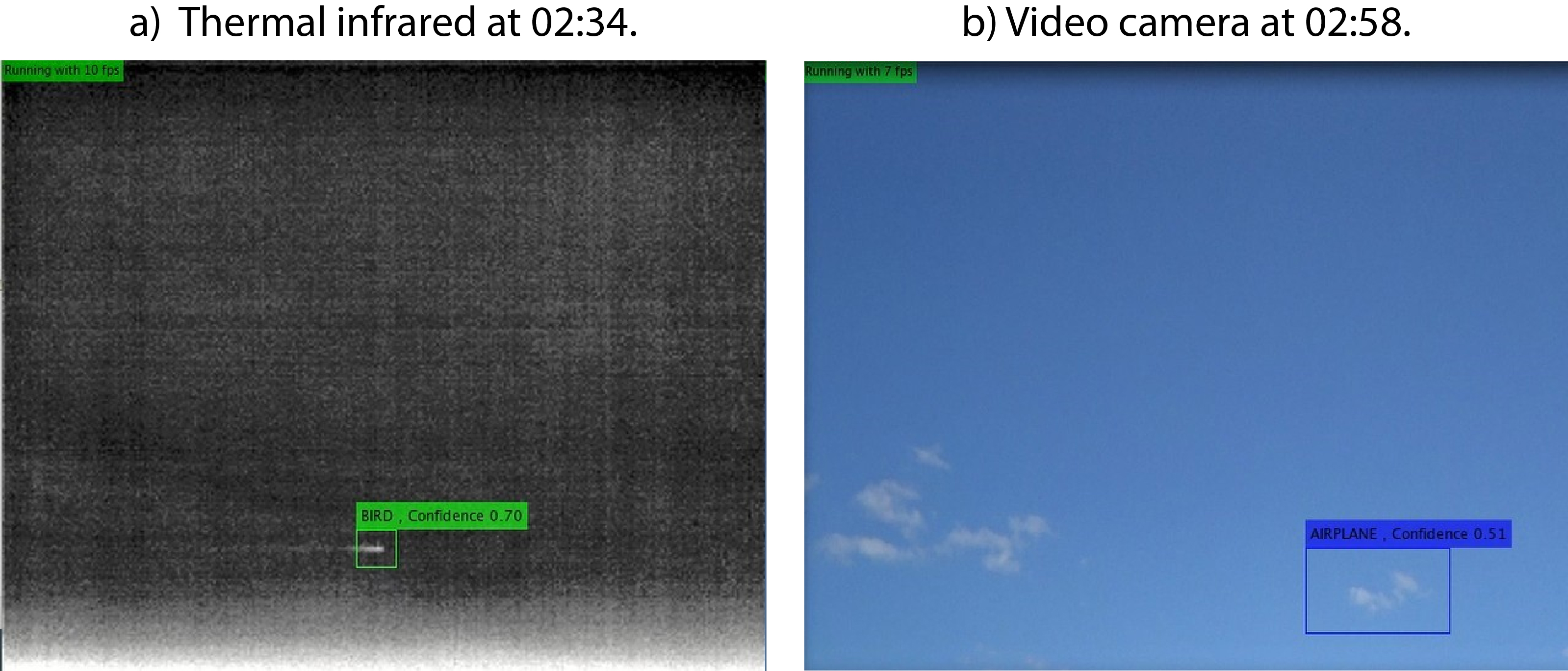}
\caption{False detections of the thermal infrared at 02:34 and the video camera at 02:58 indicated in Figure~\ref{fig:fusion-results1}b. Pictures originally appearing in \cite{drone20thesis}.}
\label{fig:fig43}
\end{figure}

We have also looked at the system behaviour without a drone flying in front of it to analyse false detections.
To do this, a ten-minute section of videos from the evaluation sessions was reviewed frame-by-frame.
Figure~\ref{fig:fusion-results1}b shows the timestamps, sensor types, causes of false detection, and resulting output labels. 
Setting the minimum number of sensors option to two prevents all the false detections in the figure from becoming false detections on a system level.
The false detections caused by insects flying just in front of the sensors are very short-lived, while the ones caused by clouds can last longer, up to several seconds.
Figure~\ref{fig:fig43} shows the false detections of the thermal infrared at 02:34 and the video camera at 02:58.
As described earlier, the individual weaknesses observed for the primary sensors are  sensitivity to clouds (thermal infrared) and autofocus problems (video camera). However, the fusion detection pipeline shows that such individual shortcomings can be overcome with a multi-sensor solution. 

Some other screenshots from the system evaluation sessions and interesting complementary observations are also pointed out in Appendix~\ref{sect:appendix}.

%
%

\section{Conclusions}
\label{sect:conclusions}

The increased use of drones for many recreational and professional purposes is creating concerns about the safety and security of premises where such vehicles may create risky situations or compromise security. 
Here, we explore the design of a multi-sensor drone detection system that employs state-of-the-art feature extraction and machine learning techniques, e.g. YOLOv2 detector \cite{Redmon17}, GMM background subtraction \cite{Stauffer99,gmmmatlab}, Kalman filters \cite{kalmanmatlab}, MFCC audio features or LSTM classifiers \cite{Hochreiter97}.
We employ a standard video camera and audio microphone complemented with a thermal infrared camera.
These two cameras are steered towards potential objects of interest with a fish-eye lens camera with a wider field of view that is used to detect moving objects at once over a bigger portion of the sky. 
We also include an ADS-B receiver, which allows for tracking cooperative aircrafts that broadcast their vehicle type, and a GPS receiver.
We have also evaluated a radar module, which has been finally discarded due to its very short practical range.
The sensors are all placed on a pan/tilt platform mounted on a standard surveyor's tripod.
This allows easy deployment of the solution outdoors and portability since the system can be disassembled into a few large parts and placed in a transport box.

The performance of the individual sensors is evaluated in terms of precision, recall, F1-score and mean average precision (mAP).
We observe that thermal infrared sensors are suitable for the drone detection task with machine learning techniques. 
The infrared detector achieves an F1-score of 0.7601, showing similar performance as the visible video detector with an F1-score of 0.7849. The audio classifier achieves an F1-score of 0.9323.
By sensor fusion, we also make the detection and classification more robust than any of the sensors individually, showing the efficiency of sensor fusion as a means to mitigate false detections. 
Another novelty is the investigation of detection performance as a function of the sensor-to-target distance.
We also contribute with a multi-sensor dataset, which overcomes the lack of publicly available data for this task. %
Apart from the feasibility of the sensors employed, especially the thermal infrared one, our dataset also uses an expanded number of target classes compared to related papers.
Our dataset is also especially suited for the comparison of infrared and visible video detectors due to the similarities in conditions and target types in the set.
To the best of our knowledge, we are also the first to explore the benefits of including ADS-B data to better separate targets prone to be mistaken for drones.

Future research could implement a distance estimation function based on the output from the detectors. Such research could also include investigating distributed detection and tracking and further using the temporal dimension to separate drones from other targets based on their behaviour over time as the system tracks them.
In a distributed scenario \cite{Yin20ojsp_cooperative}, several agents (detection stations in our case) could cooperate, each having access to only a few sensing modalities or varying computing capabilities or battery power.
Replicating one collection station with many different sensors may be cost-prohibitive or difficult if transport and deployment need to be done quickly (e.g. we used a lighter version with only two sensors and manual steering for dataset collection, Figure~\ref{fig:data_collection}). 
However, cooperative operation entails challenges, such as handling orchestration and task distribution between units, connectivity via 4G/5G networks, or cloud/edge computing if the deployed hardware lacks such capability. 
Another challenge is the optimal fusion of information from sources that may have different data quality (e.g. sensors with different features), different modalities (e.g. visible, NIR, audio...), or different availability at a given time (e.g. a specific sensor is not deployed or the object is not visible from that location). In the case of visual sensors, there is also a need to match targets observed by sensors in different physical locations since the object is seen from a different point of view.  On the other hand, the first station that detects a target can notify it and provide helpful information to aid in the detection by the others.  
Another direction is to implement the YOLO detector in the fish-eye camera. However, this would demand a training dataset or skewing video camera images to match the fish-eye lens distortion.
It would also be interesting to use YOLO v3 instead since it is more efficient in detecting small objects according to \cite{Unlu19}, or even more recent versions, which currently go up to YOLO v7.
In this paper, we kept using YOLO v2 since it is the predominant choice in the literature (as seen in recent surveys \cite{9765451}), enabling a fairer comparison of our results with previous works.
The performance of the audio classifier with respect to the sensor-to-target distance could also be explored in the same way as the video sensors. 
Furthermore, the influence of covariates given by different weather conditions could also be investigated.

A radar with a good range would have contributed significantly to the system results since it is the only available sensor that can measure the distance to the target efficiently. 
%
%
Likewise, it would be interesting to test the system against a drone equipped with a transponder to see the performance of the ADS-B worker at a short distance. Such transponders weigh as little as 20 grams \cite{uAvioni}. However, the price of such equipment ($\sim$3k€) is still an issue for non-professional drone users.
Since ADS-B information is included in the system, this could also be implemented as ground truth for an online learning feature. Images of ADS-B targets could be saved, and the detectors would be retrained at regular intervals using these additional images. To further enlarge the training data set, all images of detected objects could be saved and annotated manually at a later stage.

This work can also be helpful in other areas.
One example is road traffic surveillance, since most parts and scripts (except the ADS-B receiver) are applicable after appropriate retraining to detect and track vulnerable road users (pedestrians), vehicles, etc. and even specific vehicle types such as light ones (bikes, motorcycles) or heavy ones (trucks).
Another application involving the detection of people is surveillance of large areas where access is restricted, and operator monitoring can be very resource-consuming if not automated, such as outdoors.
%

%

\vspace{6pt} 



\authorcontributions{
%
This work has been carried out by Fredrik Svanström under the supervision of Fernando Alonso-Fernandez and Cristofer Englund in the context of his Master's Thesis at Halmstad University (Master's Programme in Embedded and Intelligent Systems). The thesis is available at \cite{drone20thesis}.  
Fredrik Svanström: Conceptualization, Methodology, Investigation, Data curation, Writing –review \& editing; Fernando Alonso-Fernandez: Conceptualization, Supervision, Funding acquisition, Writing – original draft; Cristofer Englund: Conceptualization, Supervision, Writing – review \& editing.
}

\funding{
%
Author F. A.-F. thanks the Swedish Research Council (VR) for funding his research.
Authors F. A.-F. and C. E. thank the Swedish Innovation Agency (VINNOVA) for funding their research.
}

\dataavailability{
%
The data used in this paper is fully described in \cite{DiBdataset} and publicly available at \cite{svanstrom_fredrik_2020_5500576}.
} 


\conflictsofinterest{The authors declare no conflict of interest.}



\abbreviations{Abbreviations}{
The following abbreviations are used in this manuscript:\\

\noindent 
\begin{tabular}{@{}ll}
MDPI & Multidisciplinary Digital Publishing Institute\\
DOAJ & Directory of open access journals\\
TLA & Three letters acronym\\
LD & Linear dichroism
\end{tabular}
}

\appendixtitles{yes} 
\appendixstart
\appendix
\section[\appendixname~\thesection]{Complementary screenshots and observations to the sensor fusion evaluation sessions}
\label{sect:appendix}



In this section, some screenshots from the system evaluation sessions and some interesting observations are pointed out.
The images also indicate the FPS performance. 
The system can process 6 FPS or more from all cameras and over 10 processing cycles per second for the input audio stream. 
To be able to evaluate the system performance, a screen recording software has also been running on the computer at the same time as the system software. 
%
%

\begin{figure} [htb]
\centering
\includegraphics[width=0.9\textwidth]{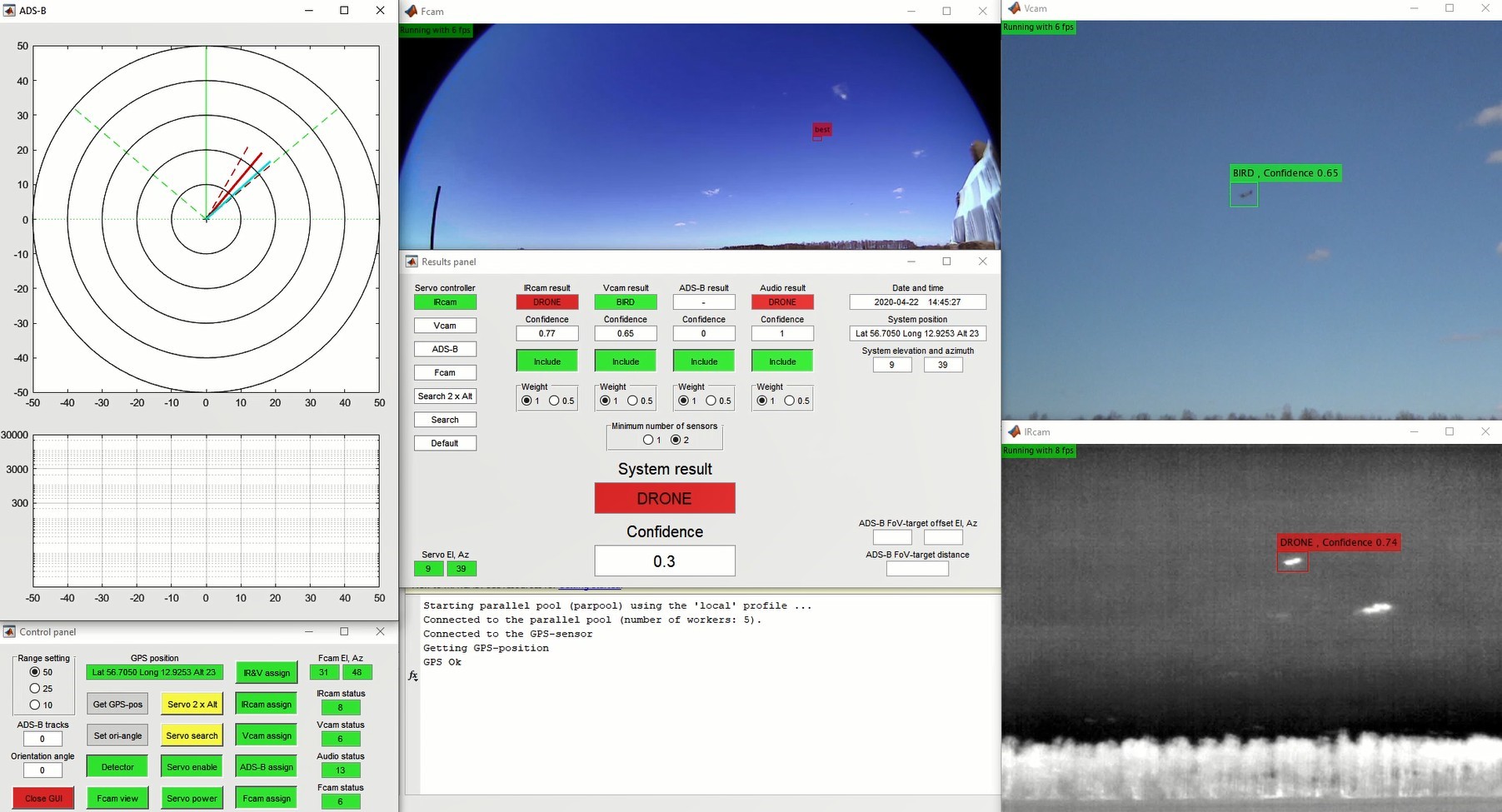}
\caption{A correct system output even if the video camera classifies the drone as a bird. Picture originally appearing in \cite{drone20thesis}.}
\label{fig:fig44}
\end{figure}

\begin{figure} [htb]
\centering
\includegraphics[width=0.9\textwidth]{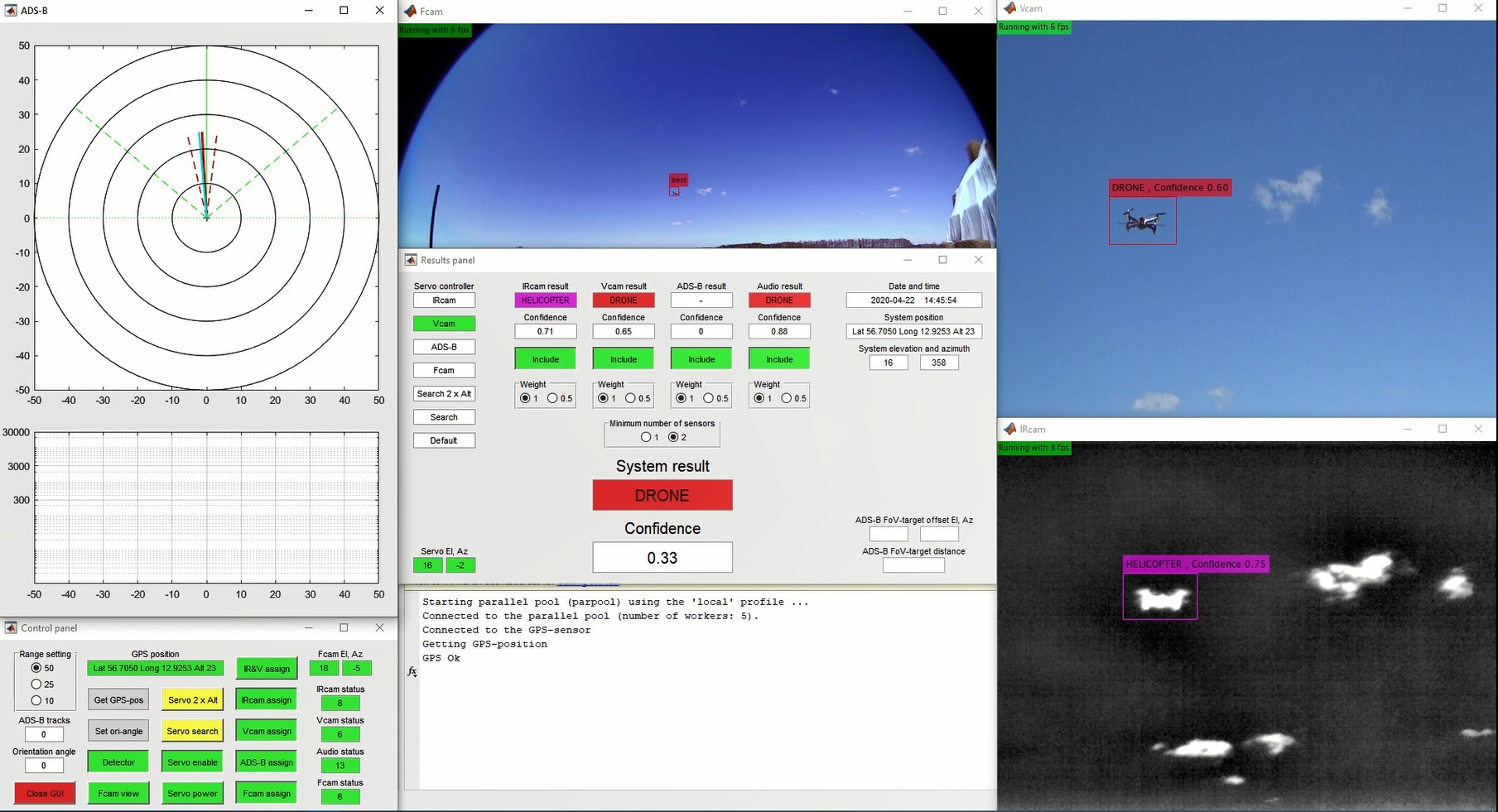}
\caption{A correct system output even if the thermal infrared camera classifies the drone as a helicopter. Picture originally appearing in \cite{drone20thesis}.}
\label{fig:fig45}
\end{figure}

The ideal situation is that all the sensors output the correct classification of the detected target and that the fish-eye camera tracks the object. This is, however, far from the case at all times.
Nevertheless, after the sensor fusion, the system output class is observed to be robust, as shown in Figure~\ref{fig:fig44} and \ref{fig:fig45}, where the video and thermal infrared cameras classify the drone incorrectly, but still with a correct system output.

\begin{figure} [htb]
\centering
\includegraphics[width=0.9\textwidth]{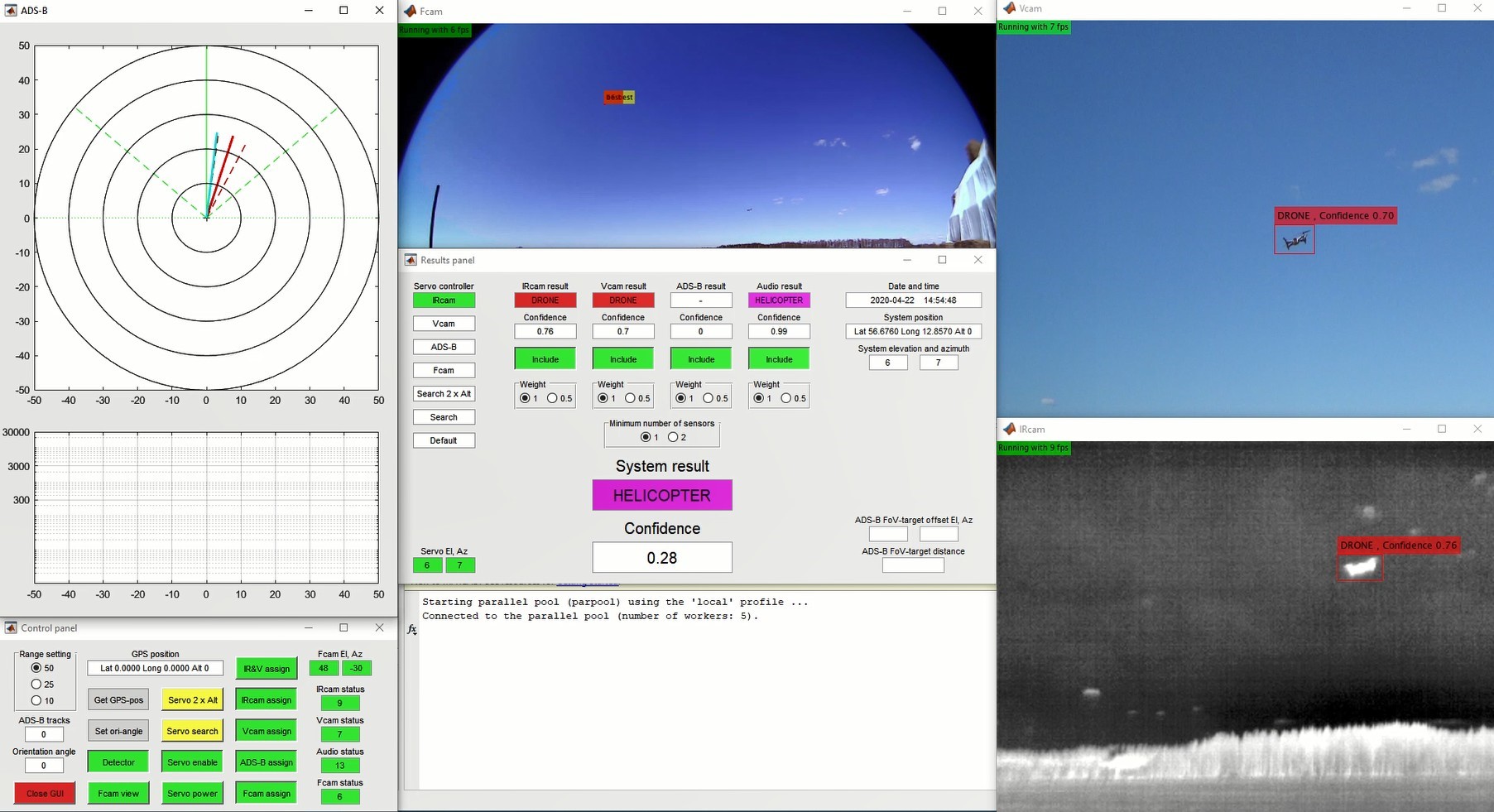}
\caption{The high confidence score of the audio classifier causes the system output to be incorrect, just like the audio output. Picture originally appearing in \cite{drone20thesis}.}
\label{fig:fig46}
\end{figure}

\begin{figure} [htb]
\centering
\includegraphics[width=0.9\textwidth]{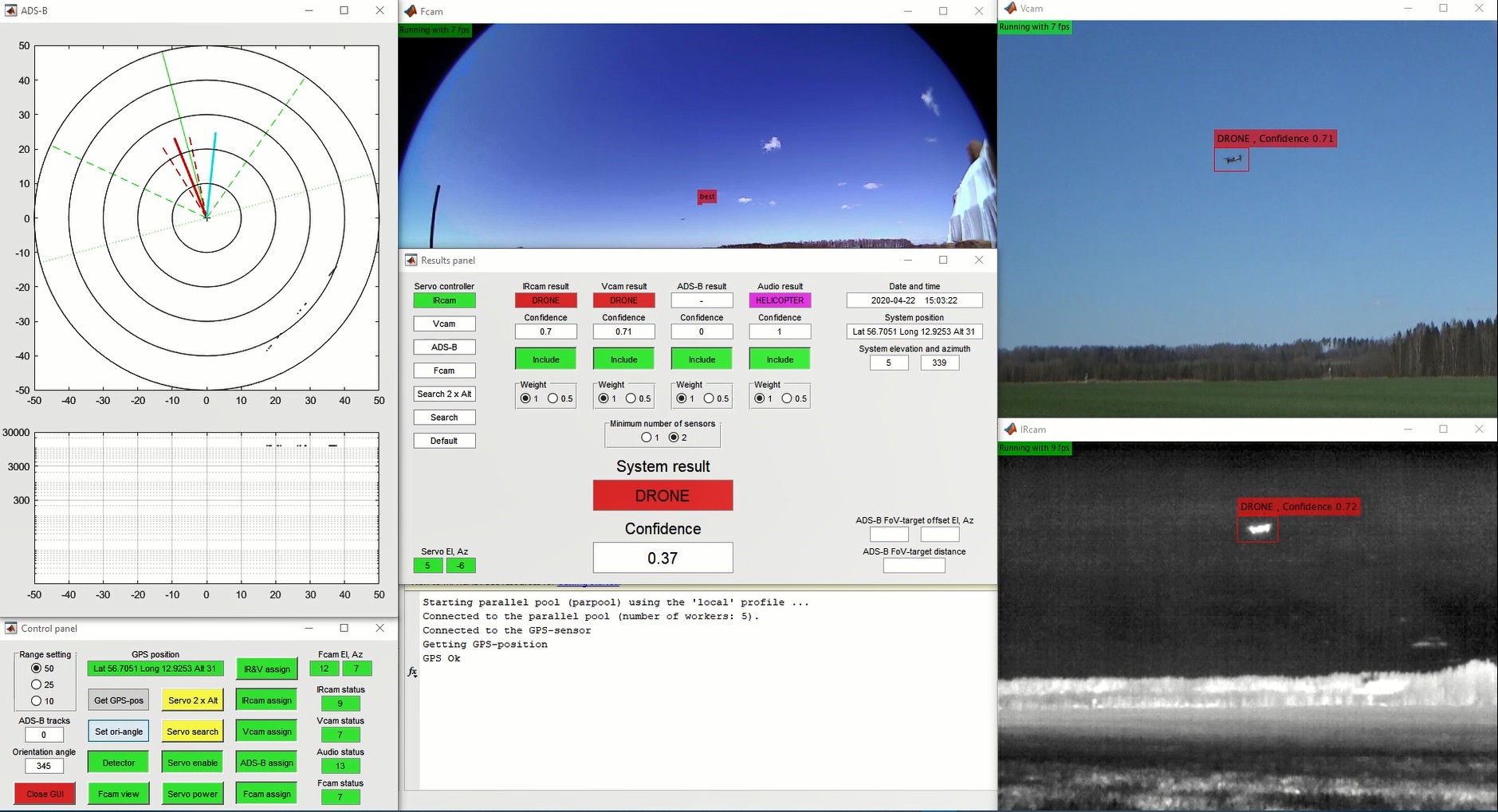}
\caption{The time smoothing part of the sensor fusion reduces the effect of an occasional misclassification, even if that has a high confidence score. Picture originally appearing in \cite{drone20thesis}.}
\label{fig:fig47}
\end{figure}

\begin{figure} [htb]
\centering
\includegraphics[width=0.9\textwidth]{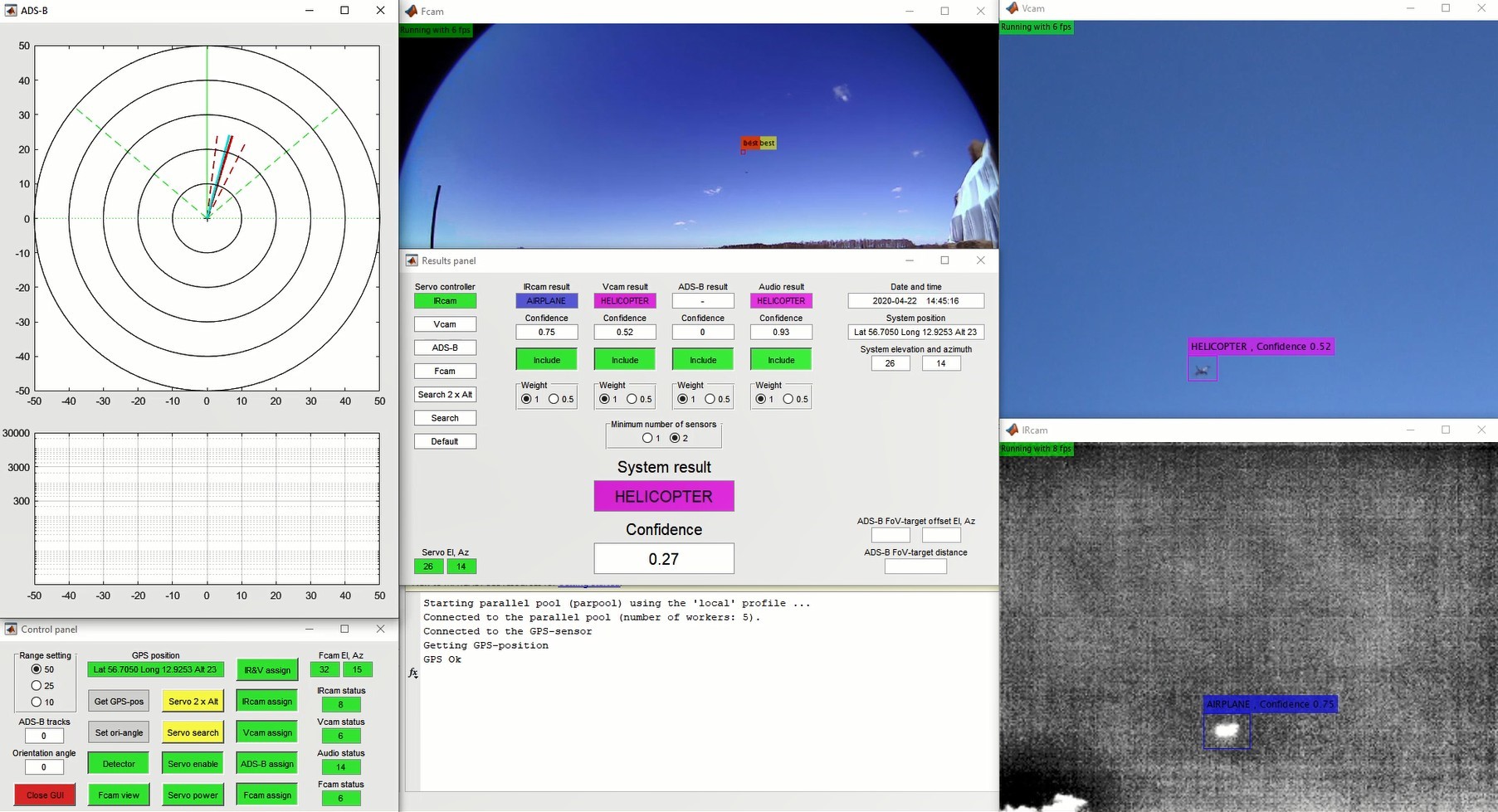}
\caption{Several sensors misclassify the drone at the same time. Picture originally appearing in \cite{drone20thesis}.}
\label{fig:fig48}
\end{figure}

Since the output class depends on the confidence score, the result is sometimes the opposite, as shown in Figure~\ref{fig:fig46}, so a very confident sensor (audio) causes the system output to be wrong. 
If this turns out to be frequent, the weight of the sensor can easily be adjusted, or the sensor can be excluded entirely from the system result. 
The time smoothing procedure of the sensor fusion will also reduce the effect of an occasional misclassification so that the system output stays correct, as seen in Figure~\ref{fig:fig47}. Naturally, there are also times when all sensors are wrong, as evident in Figure~\ref{fig:fig48}.

\begin{figure} [htb]
\centering
\includegraphics[width=0.9\textwidth]{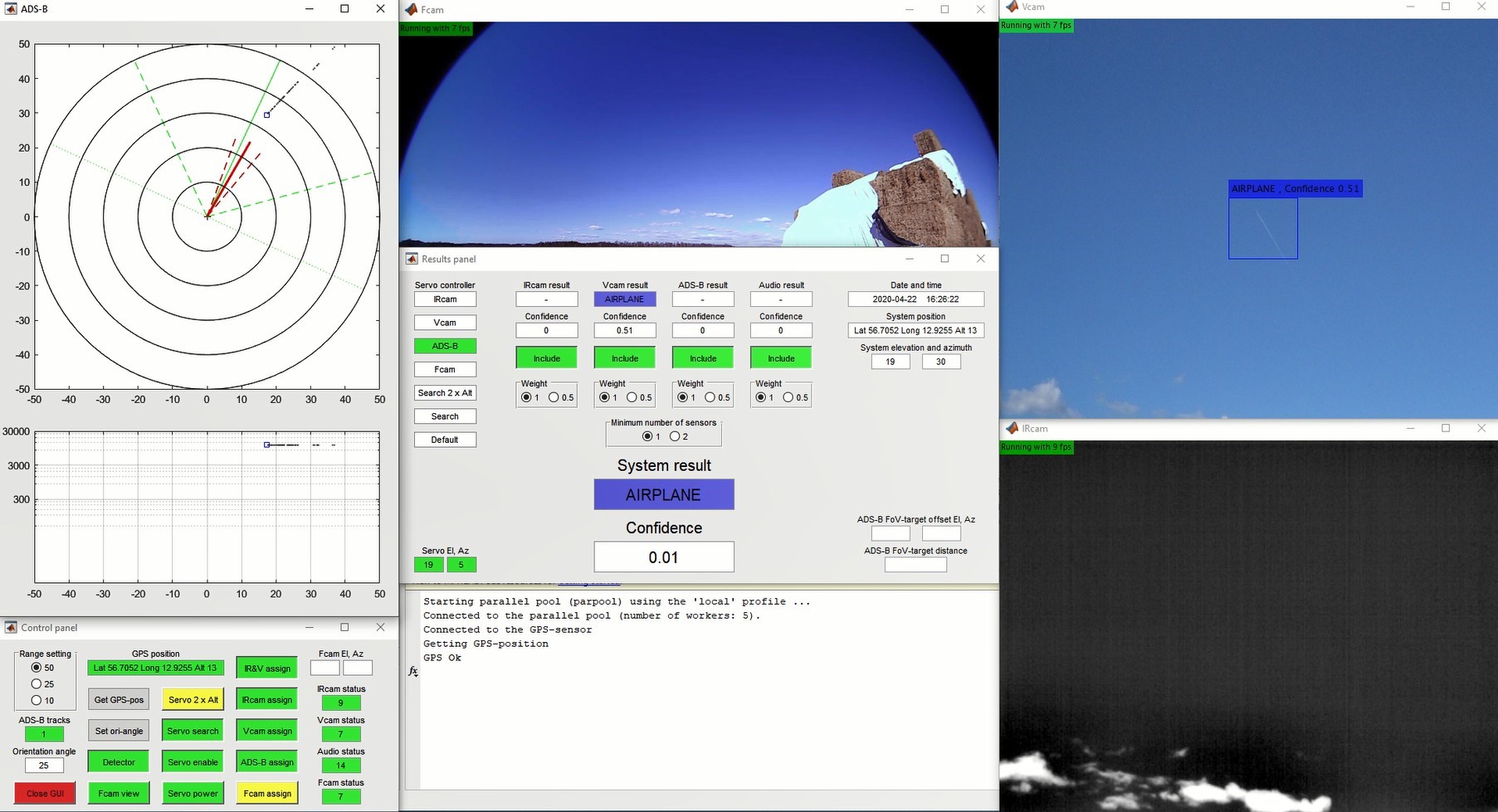}
\caption{An airplane detected and classified correctly by the video camera worker at a sloping distance of more than 35000 m. See the text for details. Picture originally appearing in \cite{drone20thesis}.}
\label{fig:fig32}
\end{figure}

To the best of our knowledge, the inclusion of an ADS-B receiver in a drone detection system has not yet been described in the scientific literature, so it is interesting to see how this information is utilized.
In Figure~\ref{fig:fig32}, an airplane is detected and classified at a sloping distance of more than 35000 m. The ADS-B information guides the pan/tilt platform in the direction of the airplane so that the video camera can detect it.
At this distance, even a large commercial airplane is only about 1.4 pixels, hence well below the detection limit of the DRI criteria.
The reason behind the fact that the ADS-B FoV-target distance display is empty is that the main script will not present that information until the target is within a 30000 m horizontal distance. This limit is set based on the assumption that no target beyond 30000 m should be detectable, which turned out to be wrong.

\begin{figure} [htb]
\centering
\includegraphics[width=0.9\textwidth]{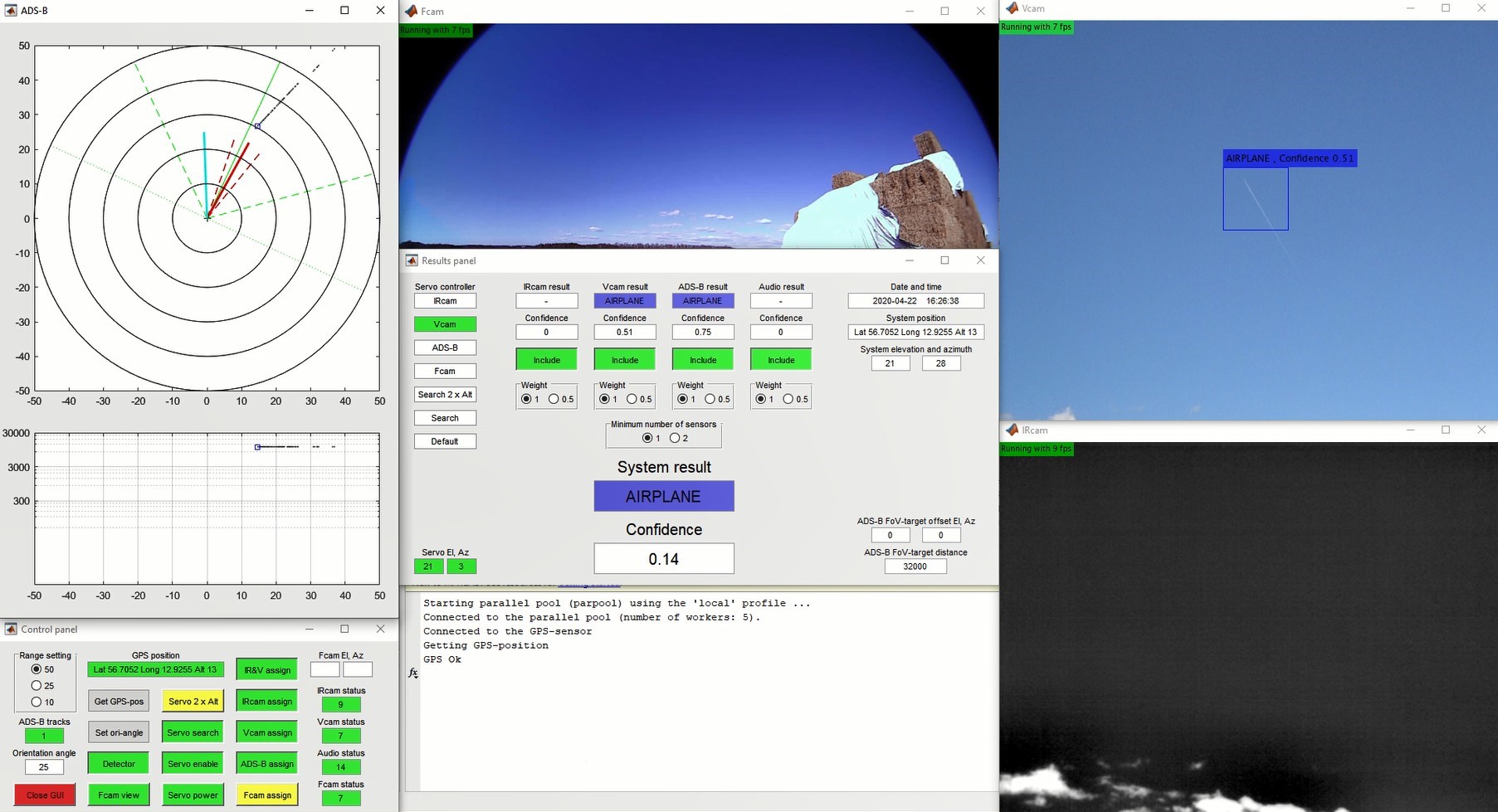}
\caption{When the airplane is within 30000 m horizontal distance the ADS-B information is presented in the results panel. Picture originally appearing in \cite{drone20thesis}.}
\label{fig:fig49}
\end{figure}

\begin{figure} [htb]
\centering
\includegraphics[width=0.9\textwidth]{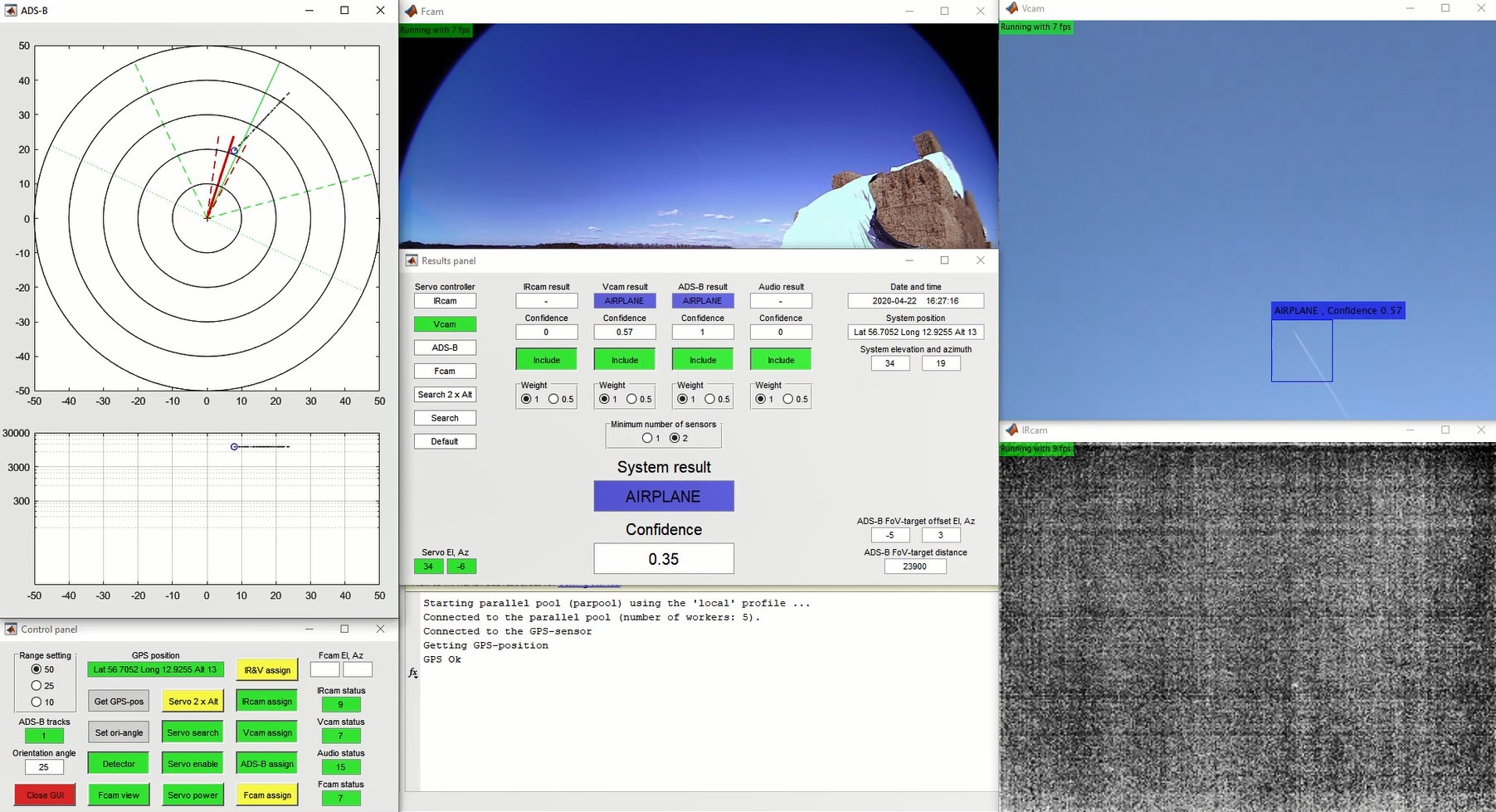}
\caption{The system has received the vehicle category message, so the confidence for the airplane classification is set to 1. The airplane is now at a distance of 23900m. Picture originally appearing in \cite{drone20thesis}.}
\label{fig:fig50}
\end{figure}

\begin{figure} [htb]
\centering
\includegraphics[width=0.9\textwidth]{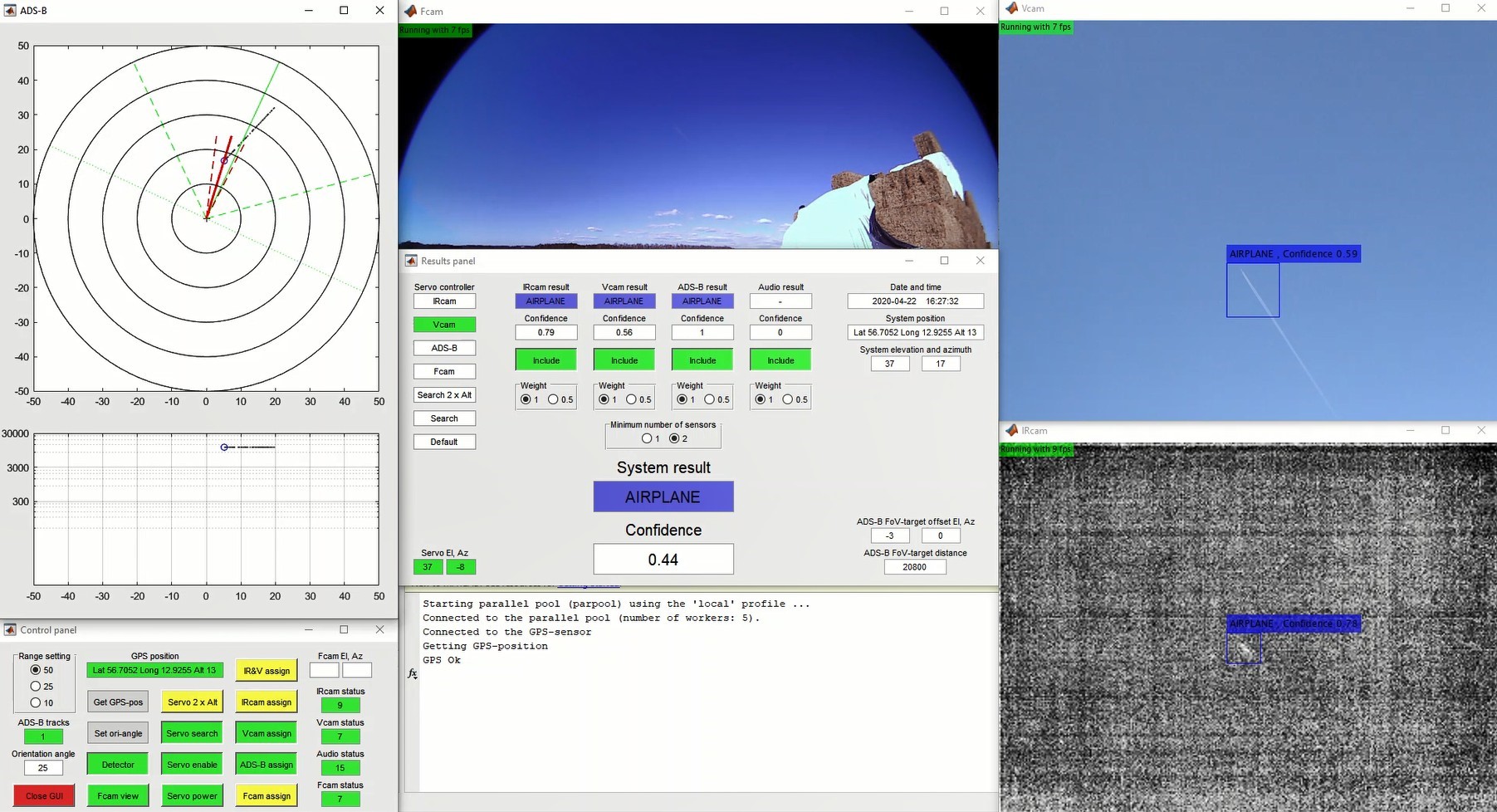}
\caption{The thermal infrared worker detects the airplane at a distance of 20800 m. Picture originally appearing in \cite{drone20thesis}.}
\label{fig:fig51}
\end{figure}

Looking at Figure~\ref{fig:fig49}, we can see that the ADS-B information will appear in the results panel when the airplane comes within 30000 m horizontal distance from the system. At this moment, the sloping distance is 32000 m, and the offset between the camera direction and the calculated one is zero. Moreover, since the system has not yet received the vehicle category information, the target is marked with a square in the ADS-B presentation area, and the confidence score of the ADS-B result is 0.75.
The next interesting event, shown in Figure~\ref{fig:fig50}, is when the system receives the vehicle category message. To indicate this, the symbol in the ADS-B presentation is changed to a circle, and the confidence is set to one since we are sure that it is an airplane. At a distance of 20800 m, it is also detected and classified correctly by the thermal infrared worker, as shown in Figure~\ref{fig:fig51}.

\clearpage

\begin{adjustwidth}{-\extralength}{0cm}

\reftitle{References}



\bibliography{bibliography}

\end{adjustwidth}
\end{document}